\RequirePackage[svgnames]{xcolor}
\documentclass[11pt,letterpaper]{mystyle}
\usepackage{amsmath,amsfonts,bm}

\def\eqref#1{equation~\ref{#1}}
\def\1{\bm{1}}

\DeclareMathAlphabet{\mathsfit}{\encodingdefault}{\sfdefault}{m}{sl}
\SetMathAlphabet{\mathsfit}{bold}{\encodingdefault}{\sfdefault}{bx}{n}

\usepackage[comma,authoryear,compress]{natbib}
\usepackage{adjustbox}
\usepackage{multirow}
\usepackage{array}
\usepackage{makecell}
\usepackage{wrapfig}
\usepackage{placeins}
\usepackage{float}
\usepackage{xspace}
\usepackage[scaled=0.94]{helvet}
\usepackage{etoolbox}
\usepackage[all]{hypcap}

\hypersetup{
  colorlinks=true,
  linkcolor=YaleBlue,
  citecolor=YaleBlue,
  urlcolor=YaleBlue,
  pdftitle={Look Beyond the Question: Preserve--Focus--Fuse for Whole-Slide Image Question Answering},
  pdfauthor={Xiangyu Li, Jincheng Li, Tianle Du, Yanheng Li, Haoxuan Che, Yuxin Sun, Long Bai, Liangliang Liu, Ran Su}
}

\title{\makebox[\linewidth][c]{Look Beyond the Question: Preserve--Focus--Fuse}\\
\makebox[\linewidth][c]{for Whole-Slide Image Question Answering}}
\runningtitle{Look Beyond the Question: Preserve--Focus--Fuse for WSI Question Answering}

\makeatletter
\def\@author{%
  {\centering\normalfont\fontsize{11.2}{17}\selectfont
  \makebox[\linewidth][c]{%
    \mbox{Xiangyu Li\textsuperscript{1,2,*}}\quad
    \mbox{Jincheng Li\textsuperscript{2,3,4,*}}\quad
    \mbox{Tianle Du\textsuperscript{1}}\quad
    \mbox{Yanheng Li\textsuperscript{5}}\quad
    \mbox{Haoxuan Che\textsuperscript{6}}}\\
  \makebox[\linewidth][c]{%
    \mbox{Yuxin Sun\textsuperscript{4}}\quad
    \mbox{Long Bai\textsuperscript{7}}\quad
    \mbox{Liangliang Liu\textsuperscript{8}}\quad
    \mbox{Ran Su\textsuperscript{1,$\dagger$}}}\par}
  \vspace{8pt}
  {\centering\normalfont\fontsize{9}{13}\selectfont
  \mbox{\textsuperscript{1}Tianjin University},\enspace\allowbreak
  \mbox{\textsuperscript{2}Shanghai AI Laboratory},\enspace\allowbreak
  \mbox{\textsuperscript{3}University of Science and Technology of China},\enspace\allowbreak
  \mbox{\textsuperscript{4}Nantong University},\enspace\allowbreak
  \mbox{\textsuperscript{5}City University of Hong Kong},\enspace\allowbreak
  \mbox{\textsuperscript{6}The Hong Kong University of Science and Technology},\enspace\allowbreak
  \mbox{\textsuperscript{7}The Chinese University of Hong Kong},\enspace\allowbreak
  \mbox{\textsuperscript{8}Henan Agricultural University}\par}
  \vspace{5pt}
  {\centering\normalfont\fontsize{8.5}{11}\selectfont
  \textsuperscript{*}Equal contribution.\quad
  \textsuperscript{$\dagger$}Corresponding author.\par}%
}
\makeatother
\definecolor{AbstractBackground}{HTML}{F0F4F7}
\definecolor{FrontRule}{HTML}{647D91}
\definecolor{USTCBlue}{cmyk}{1,0.8,0,0}
\renewcommand{\headingfont}{\color{YaleBlue}\bfseries\fontsize{13}{14}\selectfont}

\newcommand{\institutionwordmark}[2]{%
  \raisebox{-.5\height}{\includegraphics[width=#2]{logos/#1}}}
\newcommand{\institutionbanner}{%
  {\centering
  \makebox[\linewidth][s]{%
    \institutionwordmark{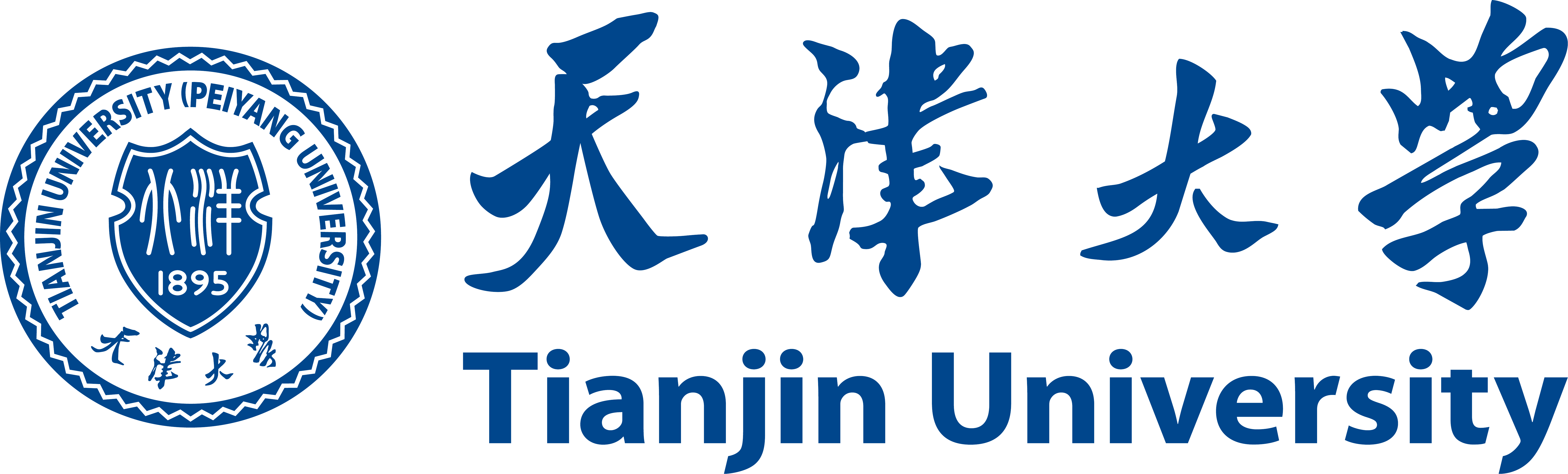}{99pt}\hfill
    \institutionwordmark{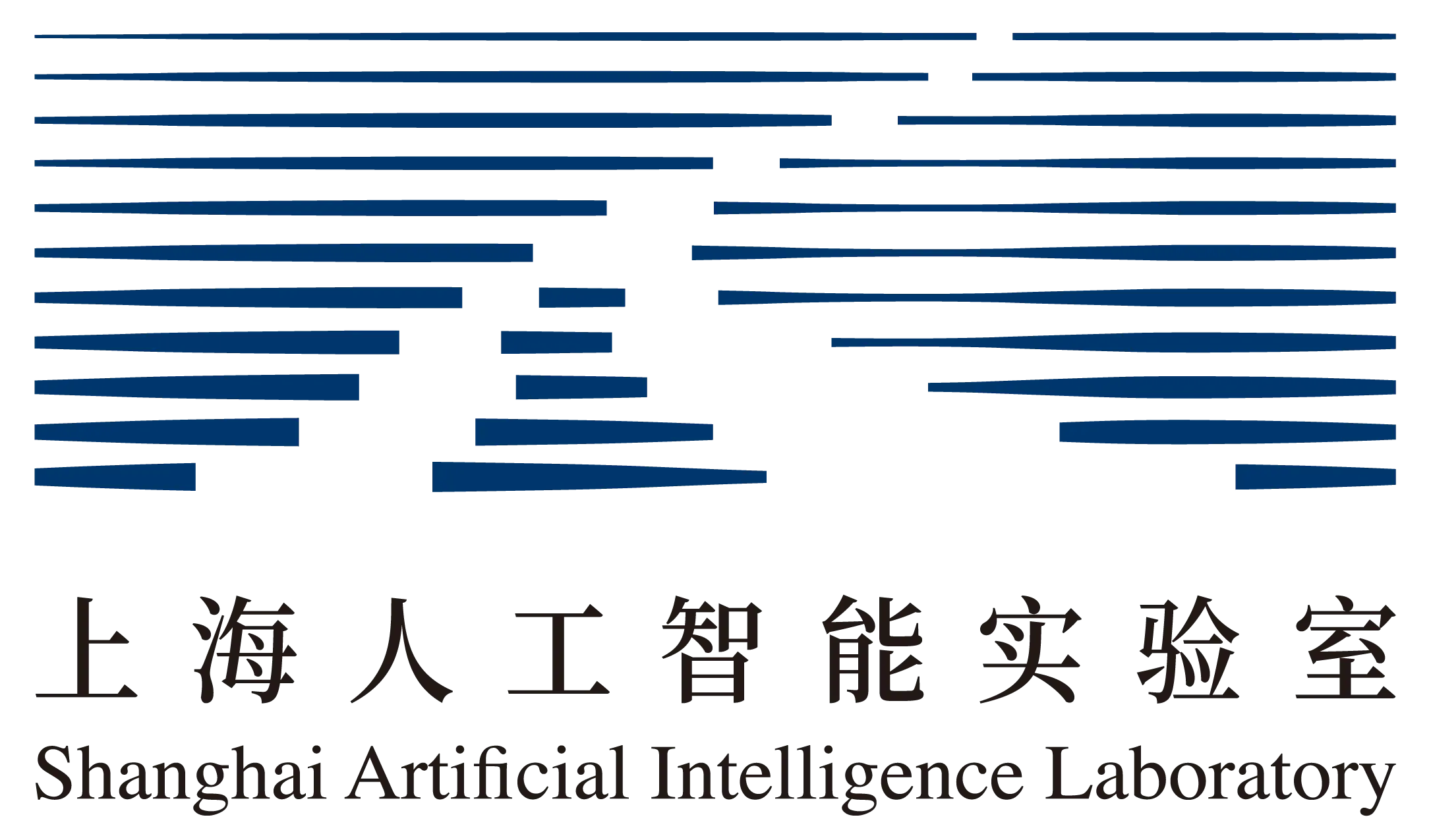}{65pt}\hfill
    \textcolor{USTCBlue}{\institutionwordmark{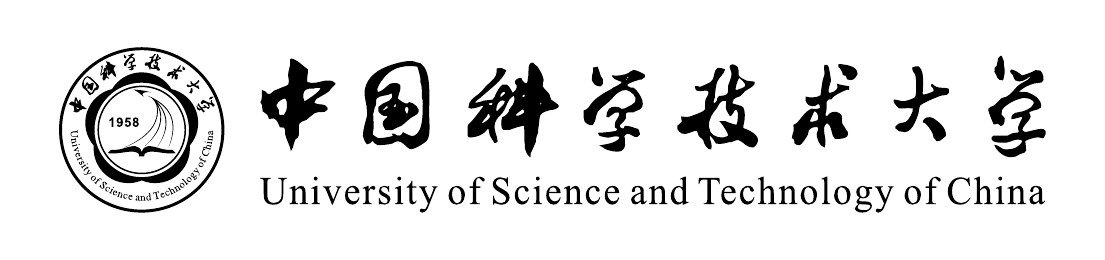}{127pt}}\hfill
    \institutionwordmark{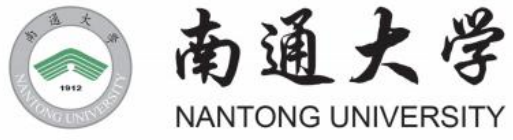}{100pt}}\par
  \vspace{13pt}
  \makebox[\linewidth][s]{%
    \institutionwordmark{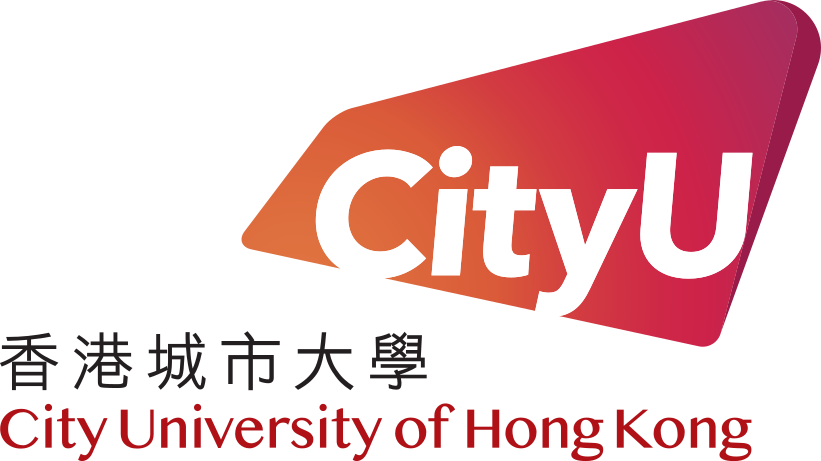}{69pt}\hfill
    \institutionwordmark{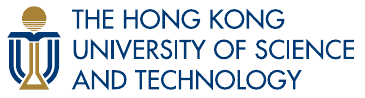}{120pt}\hfill
    \institutionwordmark{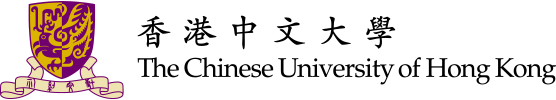}{127pt}\hfill
    \raisebox{-.5\height}{\adjustbox{width=126pt}{%
      \clipbox{0pt 0pt {0.5\width} 0pt}{\includegraphics{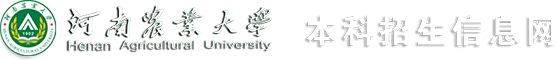}}}}}\par}%
}

\fancypagestyle{coverpage}{%
  \fancyhf{}

}

\makeatletter
\renewcommand{\maketitle}{%
  \thispagestyle{coverpage}
  \vspace*{-12mm}
  {\color{FrontRule}\hrule height 0.5pt}\par
  \vspace{16pt}
  {\centering\normalfont\bfseries\color{black}\fontsize{22.5}{27}\selectfont
    \@title\par}
  \vspace{14pt}
  {\color{FrontRule}\hrule height 0.5pt}\par
  \vspace{16pt}
  {\@author\par}
  \vspace{18pt}
  \institutionbanner
  \vspace{20pt}
  \begin{tcolorbox}[
    colback=AbstractBackground,colframe=AbstractBackground,
    boxrule=0pt,arc=8pt,outer arc=8pt,
    left=16pt,right=16pt,top=16pt,bottom=17pt,
    before skip=0pt,after skip=0pt]
    {\centering\sffamily\bfseries\color{YaleBlue}\fontsize{14}{17}\selectfont
      Abstract\par}
    \vspace{11pt}
    {\normalfont\color{black}\fontsize{10.5}{14}\selectfont
      \setlength{\parindent}{0pt}\theabstract\par}
  \end{tcolorbox}
  \clearpage
}
\makeatother

\newcommand{\method}{\textsc{BeyondPath}\xspace}

\definecolor{BestRed}{RGB}{220,0,0}
\definecolor{SecondBlue}{RGB}{0,70,210}
\definecolor{OursRowBlue}{RGB}{232,242,252}
\newcommand{\best}[1]{\textcolor{BestRed}{\ifmmode\mathbf{#1}\else\bfseries #1\fi}}
\newcommand{\second}[1]{\textcolor{SecondBlue}{\ifmmode\mathbf{#1}\else\bfseries #1\fi}}
\newcommand{\ranklegend}{Best and second-best results are shown in \textcolor{BestRed}{red} and \textcolor{SecondBlue}{blue}, respectively.}
\newcommand{\tablefont}{\footnotesize}

\begin{document}
\begin{abstract}
A question about a whole-slide image rarely names every morphological
finding needed to answer it. Question-guided patch selection can therefore
overlook supporting morphology, while slide-level compression can dilute
focal evidence. Inspired by how pathologists combine broad slide review,
question-directed examination, and integration of their findings, we
introduce \method, a Preserve--Focus--Fuse framework for whole-slide image
question answering. Preserve learns a question-agnostic Morphology Bank
through answer supervision across multiple questions about each training
slide. Focus retrieves question-specific evidence from all foreground
patches, including regions outside the Bank. Fuse integrates the preserved
morphology at the retrieved evidence positions through
Morphology-on-Visual-Evidence (MoV) Fusion, producing a fixed 96-token visual
prefix. On WSI-Bench, \method achieves an Avg-9 of 0.769 and leads all eight
reported report-generation metrics using the same checkpoint. It also
achieves 0.639 zero-shot overall accuracy on SlideBench-BCNB. Controlled
ablations show complementary benefits from morphology preservation and
question-directed retrieval, and support cross-question supervision and
fusion at the evidence positions. These results support looking beyond the
question's explicit cues: preserving broader slide morphology while retaining
the ability to seek evidence for each question.
\end{abstract}
\maketitle

\section{Introduction}

Whole-slide image (WSI) question answering requires locating and interpreting
evidence distributed across a gigapixel tissue section. Pathology multimodal
large language models (MLLMs) have made this task increasingly accessible,
supporting natural-language questions, report generation, and interactive
slide analysis \citep{chen2024wsi,chen2025slidechat,liang2025wsi,sun2025cpath}.
Yet a question specifies what must be answered without necessarily naming
every morphological finding needed to answer it. A question about tumor
grade, for example, may leave the relevant tissue architecture and cellular
features implicit. Selecting regions that match the question can thus leave
out supporting morphology. Effective WSI-VQA must look beyond the question's
explicit cues while remaining focused on the requested decision.

\begin{figure*}[t]
    \centering
    \includegraphics[width=\linewidth]{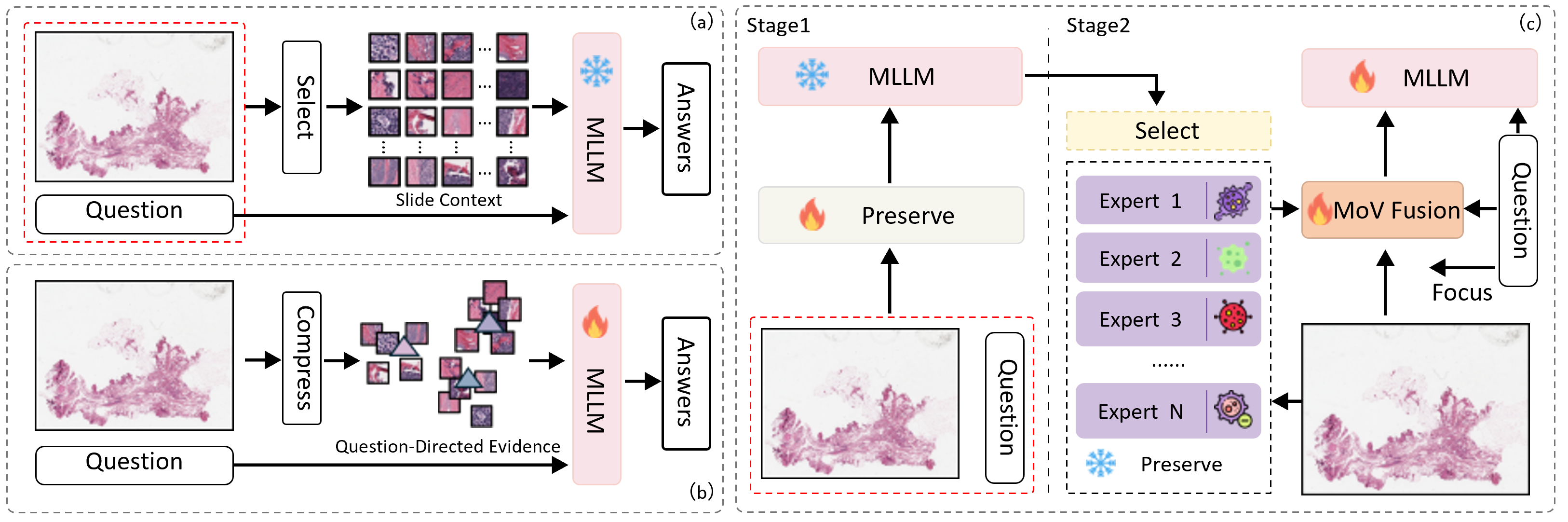}
    \caption{
    Preserve--Focus--Fuse looks beyond explicit question cues.
    (a) Question-guided selection can miss supporting morphology;
    (b) broad compression can dilute focal evidence; (c) \method preserves
    a question-agnostic Morphology Bank, retrieves evidence from all
    foreground patches, and fuses both at the retrieved positions.
    Cross-question answer supervision trains the Bank; the output is a
    96-token visual prefix. Snowflakes and flames mark frozen and trainable
    components.
    }
    \label{fig:motivation}
\end{figure*}

A single WSI can contain thousands to tens of thousands of foreground patches,
motivating long-context processing, slide encoders, pooling, resampling, and
patch selection \citep{chen2025slidechat,liang2025wsi,sun2025cpath,
chen2024wsi,huang2026act}. These approaches reduce or reorganize the patch
set so that the language model can process the slide. Broad compression can
dilute small informative regions, whereas question-conditioned selection can
omit morphology not explicitly named in the query. Moreover, restricting
question-specific retrieval to an already compressed candidate set prevents it
from recovering excluded regions. These difficulties motivate the
\emph{Relevance--Sufficiency Gap}: question-matched patches may provide an
incomplete set of findings for answering the question
(Figure~\ref{fig:motivation}(a--b)).

Pathological slide review offers a useful design cue. Broad inspection forms
an initial morphological impression; question-directed examination then seeks
pertinent findings; and the observed patterns are brought together when
interpreting the slide \citep{huang2026act}. This workflow motivates
preserving morphology before narrowing attention to a particular question,
while keeping further inspection open to regions beyond those initially
selected. We translate these complementary roles into a
\emph{Preserve--Focus--Fuse} paradigm (Figure~\ref{fig:motivation}(c)).

We introduce \method to learn these operations from slide-level
question--answer supervision. \textbf{Preserve} selects morphology-bearing
patch features without receiving the current question, forming a
\emph{Morphology Bank} that represents the initial slide impression. During
training, answers to multiple questions about the same slide supervise this
shared Bank, encouraging it to retain morphology useful across different
questions. \textbf{Focus} retrieves evidence for the current question from
all foreground patches, including regions outside the Bank. \textbf{Fuse},
implemented as \emph{Morphology-on-Visual-Evidence (MoV) Fusion}, lets each
retrieved evidence token read from the Bank. The resulting representation
combines preserved morphology with focused findings in a fixed 96-token
visual prefix for answer generation.

\noindent
Experiments support the representation and its decomposition: \method reaches
0.769 Avg-9 versus 0.754 for WSI-LLaVA, leads report generation, and obtains
0.639 zero-shot accuracy on SlideBench-BCNB. Cross-question supervision
raises Avg-9 from 0.736 to 0.769, a wrong-slide Bank lowers it to 0.677, and
MoV exceeds both concatenate--resample (0.727) and 192-token concatenation
(0.739), supporting complementary morphology and evidence-conditioned fusion.

\begin{itemize}
    \item We identify the need to preserve morphology beyond a question's
    explicit cues and formulate a Preserve--Focus--Fuse paradigm inspired
    by pathological slide review.
    \item We introduce \method, which learns a shared Morphology Bank
    through cross-question answer supervision, retrieves question-specific
    evidence from all foreground patches, and fuses both at 96 evidence
    positions through MoV Fusion.
    \item We evaluate WSI-Bench question answering and report generation,
    together with zero-shot transfer to SlideBench-BCNB. Controlled
    comparisons establish the benefits of the two streams, cross-question
    supervision, and evidence-position fusion within the same backbone.
\end{itemize}

\section{Related Work}

\textbf{Pathology vision--language modeling.}
Pathology vision--language pretraining provides the image--text alignment used
by modern pathology assistants. PLIP and CONCH learn transferable
representations from histology images and biomedical text
\citep{huang2023visual,lu2024visual}. LLaVA-Med adapts visual instruction
following to biomedical images \citep{li2023llava}, while Quilt-LLaVA uses
localized narratives from histopathology videos to construct instruction data
\citep{seyfioglu2024quilt}. These advances improve representation learning and
language interaction, but WSI-VQA also requires organizing slide-level
morphological evidence for each question.

\textbf{Whole-slide multimodal reasoning.}
WSI-VQA models patch relationships and use patch--word co-attention to
generate answers \citep{chen2024wsi}. SlideChat encodes long patch sequences
with LongNet \citep{chen2025slidechat}, WSI-LLaVA aligns global slide
representations with text \citep{liang2025wsi}, and CPath-Omni combines
multiscale regional pooling with learned query-token compression
\citep{sun2025cpath}. HistoSelect draws inspiration from pathological slide
review and retrieves evidence through a question-guided group sampler followed
by patch selection within selected tissue regions \citep{huang2026act}.
\method learns a separate Morphology Bank from answers across a slide's
training questions. Its question-conditioned retriever scores the full
foreground set, and MoV integrates the preserved morphology at the retrieved
evidence positions. This design retains both a broad morphological impression
and access to findings beyond the Bank.

\textbf{Visual token compression and selection.}
Visual token reduction considers attention, redundancy, diversity, and
instruction relevance when determining what reaches the language model.
FastV prunes tokens using language-model attention scores
\citep{chen2024image}, PACT combines pruning with clustering-based merging
\citep{dhouib2025pact}, and DivPrune selects tokens through a diversity
objective \citep{alvar2025divprune}. CDPruner conditions subset diversity on
the instruction \citep{zhang2026beyond}, while ToDRE selects diverse tokens
after the vision encoder and removes task-irrelevant tokens within the
language model \citep{li2026todre}. These methods primarily express context
and question evidence through one reduced stream or through a staged
candidate set. Our focus is the organization of morphological evidence under
a fixed language-side token budget. Preserve learns morphology useful across
questions, Focus selects findings for each question from the full slide, and
MoV combines these two sources within the evidence sequence.

\section{Method}
\label{sec:method}

\textbf{Overview.}

\begin{figure}[t]
\centering
\includegraphics[width=0.64\linewidth]{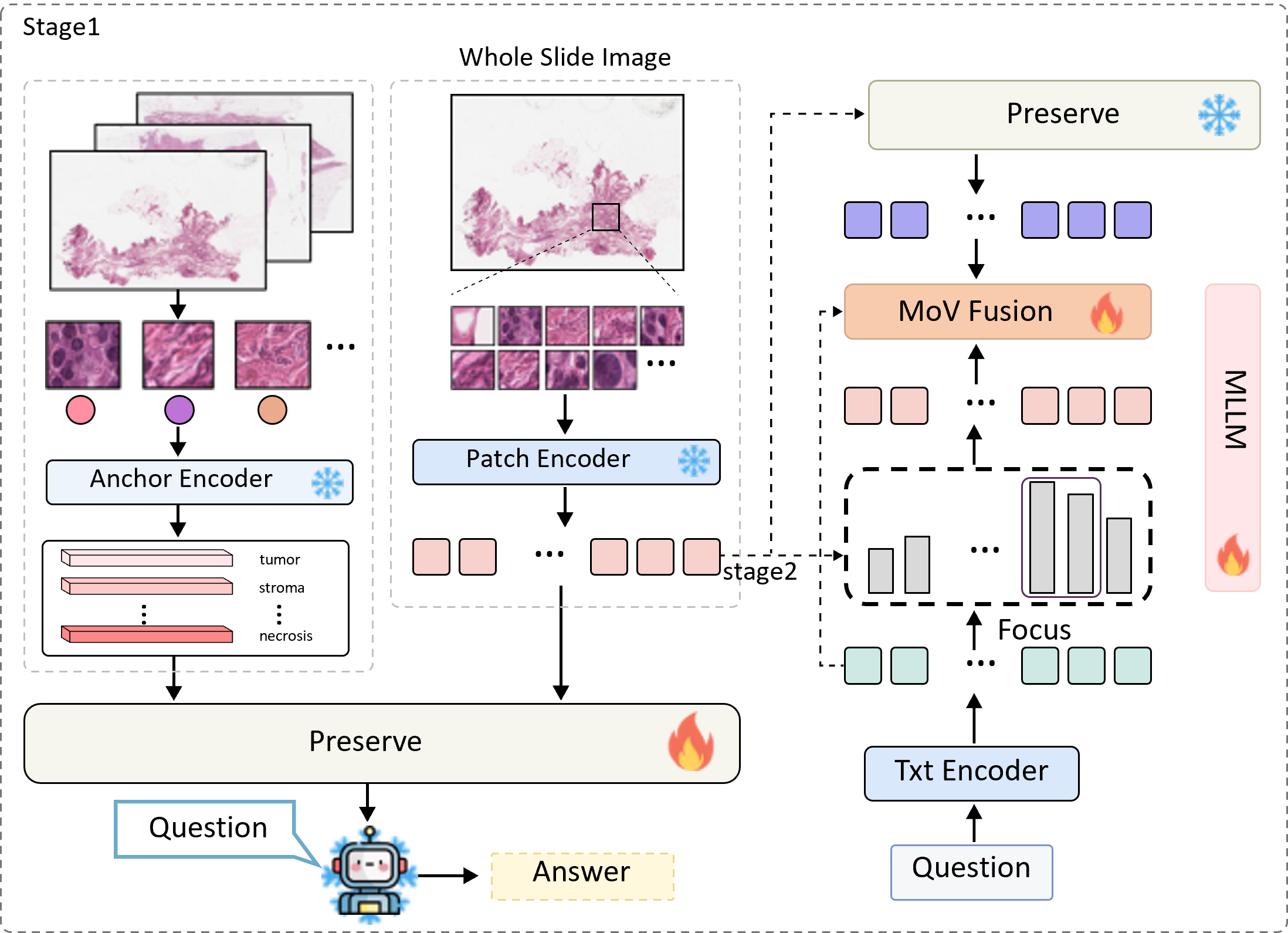}
\caption{
Overview of \method. Preserve--Focus--Fuse draws inspiration from broad slide
review, question-directed examination, and integration of findings.
Cross-question answer supervision trains Preserve to construct the Morphology
Bank. Focus retrieves evidence from all foreground patches, and MoV Fusion
integrates the Bank at the retrieved evidence positions, producing a
96-token visual prefix.
}
\label{fig:methods}
\end{figure}

Given the $s$-th WSI $S_s$, tissue filtering and regular tiling produce
$N_s$ retained foreground patches
$\mathcal{P}_s=\{p_{s,i}\}_{i=1}^{N_s}$. For a pathology question $q$,
\method constructs a fixed-length visual prefix. Stage~I learns Preserve, a
question-independent Bank constructor shared across questions for the WSI;
Stage~II freezes Preserve, retrieves current-question evidence through Focus,
and fuses it with the slide-specific Bank through MoV Fusion.
Preserve forms a question-agnostic Morphology Bank from cross-question answer
losses. Focus scores every patch in $\mathcal{P}_s$, including patches outside
the Bank. Fuse uses the retrieved evidence as queries to read the expert
banks and incorporate morphology at 96 evidence positions, so only this
evidence sequence reaches the language model.

\textbf{Features for matching and visual content.}
We retain two feature representations to distinguish pretrained
image--text matching from the visual content used in selection and fusion.
The frozen CONCH image encoder \citep{lu2024visual} produces
$f_{s,i}=\phi_v(p_{s,i})\in\mathbb{R}^{512}$ and its normalized copy
$z_{s,i}=f_{s,i}/\lVert f_{s,i}\rVert_2$.
The normalized features are used for cosine similarity in the pretrained
CONCH image--text space. An adapter maps $f_{s,i}$ to a 512-dimensional
content feature $x_{s,i}$:
\begin{equation}
x_{s,i}:=\operatorname{LN}\!\left(\operatorname{Dropout}\!\left[
\operatorname{GELU}\!\left(W_a\operatorname{LN}(f_{s,i})+b_a\right)\right]\right),
\qquad W_a\in\mathbb{R}^{512\times512}.
\label{eq:feature_adapter}
\end{equation}
Stacking the content features and normalized CONCH features row-wise gives
$X_s,Z_s^{\mathrm{vl}}\in\mathbb{R}^{N_s\times512}$, respectively.
The frozen CONCH text encoder produces the normalized question embedding
$q_c=\phi_t(q)/\lVert\phi_t(q)\rVert_2$.
$Z_s^{\mathrm{vl}}$ supports semantic anchoring and, together with $q_c$,
question--patch matching; $X_s$ supplies content for expert transformations,
evidence retrieval, and fusion. The adapter is trained in Stage~I and frozen
in Stage~II. Patch indices and slide coordinates are retained for recovering
selected regions and for visualization; absolute coordinates are not encoded
into the model.

\textbf{Bank, evidence, and prefix lengths.}
Preserve constructs expert banks
$\mathcal{B}_s=\{B_{s,e}\}_{e=1}^{N_{\mathrm{exp}}}$, with
$B_{s,e}\in\mathbb{R}^{k_e\times512}$.
Their concatenation $M_s\in\mathbb{R}^{K_m\times512}$ has length
$K_m=\sum_{e=1}^{N_{\mathrm{exp}}}k_e$.
Focus selects an evidence sequence
$E_{s,q}\in\mathbb{R}^{K_q\times512}$ from all retained foreground
patch embeddings, independently of the Bank indices.
MoV reads $\mathcal{B}_s$ using the evidence sequence and returns
$H_{s,q}\in\mathbb{R}^{K_q\times512}$.
A shared visual projector maps $H_{s,q}$ to the hidden space of
Qwen2.5-7B-Instruct \citep{hui2024qwen2}, where it is inserted before the
question tokens as a continuous visual prefix
\citep{li2023blip,liu2023visual}.
The default configuration uses six experts with $k_e=16$, giving
$K_m=96$, and selects $K_q=96$ evidence tokens.
In the fused interface, the $K_q$ MoV outputs form the language model's
visual prefix, while the Bank is read within MoV.

\subsection{Preserve: Question-Agnostic Morphology Selection}

\textbf{Cross-question answer supervision.}
Preserve learns a shared morphology selection that supports multiple
questions about the same training slide.
Let $\mathcal Q_s=\{(q_{s,j},Y_{s,j})\}_{j=1}^{m_s}$ denote the
question--answer pairs associated with training slide $s$.
During Stage~I, Preserve constructs one Morphology Bank $M_s$ from the slide
features without receiving a question. The same Bank is paired with each of
the $m_s$ questions for answer generation, and the answer losses are averaged
to update Preserve. We call this training strategy
\emph{cross-question answer supervision}. For a slide batch $\mathcal S$, we
optimize
\begin{equation}
\mathcal L_{\mathrm{CQ}}
=
\frac{1}{|\mathcal S|}
\sum_{s\in\mathcal S}
\frac{1}{m_s}
\sum_{j=1}^{m_s}
\ell_{\mathrm{ans}}
\!\left(
Y_{s,j};q_{s,j},\Pi_{\omega}(M_s)
\right),
\label{eq:cross_query_objective}
\end{equation}
where $\Pi_{\omega}$ is the visual projector and
$\ell_{\mathrm{ans}}$ is the autoregressive answer-generation loss.
The within-slide average gives each slide equal weight in this objective,
regardless of its number of question--answer pairs.
Questions enter Qwen's answer-generation branch and supervise Preserve
through gradients of this objective; their embeddings do not enter Preserve's
patch-scoring router. The selector is therefore question-agnostic by
construction, while its Bank learns to support different questions about the
same slide.

\textbf{Morphology experts.}
Different questions can require different tissue components, so Preserve
reserves Bank capacity for several morphology categories through
$N_{\mathrm{exp}}=6$ experts with fixed quotas.
Four semantic experts are anchored to tumor, stroma, immune infiltration,
and necrosis. Two free experts have no predefined semantic anchors, allowing
answer supervision to guide selection beyond these prescribed axes.
The semantic anchors cover selected common morphologies, while the free experts
allow answer supervision to retain features outside that vocabulary.
A shared router forms a 512-dimensional hidden state and produces a logit
for each patch--expert pair:
\begin{equation}
h^r_{s,i}=\operatorname{GELU}\!\left(W_r\operatorname{LN}(x_{s,i})+b_r\right),
\qquad
r_{s,i,e}=w_{r,e}^{\top}h^r_{s,i}+b_{r,e}.
\label{eq:morph_router}
\end{equation}
Each expert's logits are normalized over the patches of the current slide,
$g_{s,:,e}=\operatorname{Softmax}_i(r_{s,:,e}/\tau_r)$.
Normalizing over patches allows a region to receive high scores from
multiple experts: a tumor--stroma interface, for example, can contribute
to more than one morphology axis.

Fixed prompt ensembles in the frozen CONCH image--text space provide
semantic targets for the four anchored experts.
For each axis, patch--prompt similarities define a target distribution
for the corresponding router. The strongest similarities on the slide
also yield a soft presence estimate $\rho_{s,e}$.
This estimate weights the semantic routing loss and controls interpolation
with a learned null bank when prompt similarities provide weak support.
The free experts use $\rho_{s,e}=1$ and are trained through answer
supervision and a diversity regularizer.
A margin-based penalty on the mean relaxed selection distributions
discourages excessive overlap for free--free and semantic--free pairs.
Semantic--semantic pairs are excluded to allow overlap between the
prescribed morphology axes.
The prompt ensembles, presence calculation, and regularization terms are
specified in Appendix~\ref{app:exact_method}.
For the expert-subset variants, the two-semantic-expert configuration uses
tumor and stroma; the four-semantic-expert configurations additionally use
immune infiltration and necrosis.

\textbf{Sparse selection and Bank construction.}
Preserve retains explicit patch selections in the forward pass while
training the selection scores through a continuous relaxation.
For a score vector $\mathbf a$, let
$\mathbf P^{\mathrm{hard}}(\mathbf a,K)$ be the ordered hard Top-$K$
matrix and $\mathbf P^{\mathrm{soft}}(\mathbf a,K)$ its sequential
relaxation, which progressively downweights candidates emphasized by
earlier soft selection rows.
The straight-through selection matrix is
\begin{equation}
\mathbf P^{\mathrm{ST}}(\mathbf a,K)
=
\mathbf P^{\mathrm{soft}}(\mathbf a,K)
+
\operatorname{stopgrad}
\left(
\mathbf P^{\mathrm{hard}}(\mathbf a,K)
-
\mathbf P^{\mathrm{soft}}(\mathbf a,K)
\right).
\label{eq:st_topk}
\end{equation}
Its forward value is the hard selection matrix; gradients with respect to
the selection scores follow the relaxed matrix.
This operator gathers selected patch features in the forward pass and
provides a surrogate gradient for learning their membership.
The full relaxation is given in Appendix~\ref{app:relaxed_topk}.
With $k_e=16$ slots per expert, the selected sequence is
$V_{s,e}=\mathbf P^{\mathrm{ST}}(r_{s,:,e},k_e)X_s$.
Stage~I uses this straight-through operator; Stage~II and inference use
hard Top-$K$ for the frozen Preserve network.

Expert-specific transformations give each selected patch a representation
associated with the expert that selected it.
Each expert applies a residual $512\rightarrow1024\rightarrow512$
feed-forward network,
\begin{equation*}
\operatorname{FFN}_e(u)=W_{2,e}\operatorname{GELU}(W_{1,e}u+b_{1,e})+b_{2,e},
\end{equation*}
and adds a learned identity embedding $\eta_e$.
For a semantic expert, the presence estimate controls interpolation with
its learned null bank $N_e$:
\begin{equation}
B_{s,e}
=
\rho_{s,e}
\left[
V_{s,e}
+
\operatorname{FFN}_e
\bigl(\operatorname{LN}(V_{s,e})\bigr)
+
\mathbf{1}_{k_e}\eta_e^{\top}
\right]
+
(1-\rho_{s,e})N_e .
\label{eq:morphology_bank}
\end{equation}
For free experts, $\rho_{s,e}=1$ and the null term vanishes.
The expert banks are concatenated into
$M_s=\operatorname{Concat}_{e=1}^{6}B_{s,e}
\in\mathbb{R}^{96\times512}$.
Fixed quotas reserve slots for every expert. If several experts select
the same patch, its expert-specific token instances remain separate.
The Bank contains expert-transformed patch features and, for
semantic experts, a presence-weighted contribution from learned null banks.

\begin{figure*}[t]
\centering
\includegraphics[width=\linewidth]{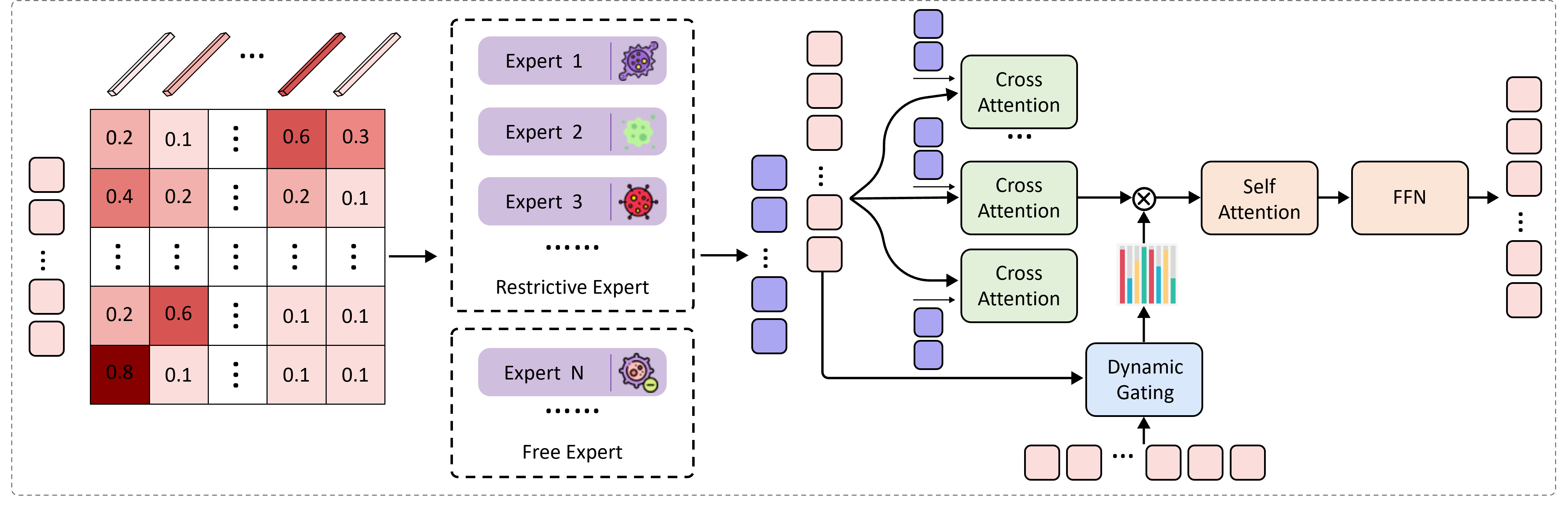}
\caption{
Preserve, Focus, and MoV implement Preserve--Focus--Fuse.
Preserve selects four semantic and two free expert banks with
relaxed-backward training followed by hard selection. Focus retrieves from
all foreground patches. MoV uses evidence tokens to query the banks and
position-specific gates while retaining 96 output positions.
}
\label{fig:module}
\end{figure*}

\subsection{Focus: Question-Conditioned Evidence Retrieval}

A Bank learned to support multiple questions can omit a finding required by a
particular question. Focus therefore scores all $N_s$ retained foreground
patch embeddings for the current question. Its candidate set includes patches
both inside and outside the Bank, and its selections may overlap with
Preserve's.

The scoring function combines pretrained image--text similarity with a
learned compatibility score.
Let $v^q_{s,i}=W_x\operatorname{LN}(x_{s,i})$ and
$\widetilde q=W_q q_c$.
With $\lambda_{\mathrm{vl}}=\operatorname{Softplus}(\beta)$, we compute
\begin{equation}
s_{s,i}
=
\lambda_{\mathrm{vl}}\,z_{s,i}^{\top}q_c
+
w_f^{\top}\tanh\!\left(v^q_{s,i}+\widetilde q+b_f\right).
\label{eq:focus_score}
\end{equation}
The first term uses cosine similarity in the frozen CONCH space; the
second learns question--patch compatibility from WSI-VQA supervision.
Focus applies the same straight-through selection operator as Preserve:
\begin{equation}
E_{s,q}
=
\mathbf P^{\mathrm{ST}}
\left(
s_{s,:},K_q
\right)X_s
+
\mathbf{1}_{K_q}\eta_{\mathrm{evi}}^{\top},
\label{eq:focus_evidence}
\end{equation}
where $\eta_{\mathrm{evi}}$ is a learned evidence-type embedding and
$K_q=96$.
The combined scores determine which patches are selected, while $X_s$
provides their visual content for fusion.
During Stage~II training, the scores receive gradients through the relaxed
selection path. At inference, Focus uses hard Top-$K$ and retains the
selected patch indices for visualization.

\subsection{Fuse: Morphology-on-Visual-Evidence Fusion}

MoV Fusion integrates the morphology retained by Preserve into the evidence
retrieved by Focus. The evidence tokens provide the queries, and the expert
banks provide the keys and values in a cross-attention interface
\citep{jaegle2021perceiver,li2023blip}. Direct concatenation would expose 192
tokens to the language model. MoV places morphology updates at the
$K_q=96$ evidence positions, which carry the fused representation into the
language model.

Starting from $H_{s,q}^{(0)}=E_{s,q}$, each MoV block performs
expert-specific cross-attention.
The current evidence states provide the queries and $B_{s,e}$ provides
the keys and values, producing
$A_{s,q,e}^{(\ell)}\in\mathbb{R}^{K_q\times512}$.
Each expert bank contains 16 tokens, so its key/value sequence length is
independent of the number of foreground patches in the slide.

\textbf{Morphology updates for each evidence position.}
Each evidence position receives its own gate weights, allowing different
anchors to draw on different expert banks.
For expert $e$, the gate concatenates the normalized evidence state,
question embedding, mean-pooled bank $\bar B_{s,e}$, expert identity
$\eta_e$, and presence estimate $\rho_{s,e}$.
A GELU MLP maps this input to an expert logit; a separate MLP produces a
null-path logit.
The gates are normalized over the six experts and a seventh path whose
attention output is zero.
Assigning weight to this null path reduces the combined contribution of
the expert-attention outputs.
The residual morphology update is
\begin{equation}
\widetilde H_{s,q,k}^{(\ell)}
=
H_{s,q,k}^{(\ell)}
+
\sum_{e=1}^{N_{\mathrm{exp}}}
\alpha_{s,q,k,e}^{(\ell)}
A_{s,q,k,e}^{(\ell)} .
\label{eq:mov_update}
\end{equation}
Each morphology update is followed by pre-normalized self-attention over
the evidence sequence and a residual SwiGLU feed-forward block.
The default model uses two 512-dimensional MoV blocks with eight attention
heads. Only the evidence positions are propagated between blocks, keeping
the output length at 96.
Each output position is initialized from one Focus-selected patch and is
then contextualized by Bank attention and interactions among evidence
tokens.

\subsection{Two-Stage Learning and Inference}

The two-stage schedule first learns the initial slide impression through
cross-question answer supervision, then learns question-conditioned retrieval
and fusion while holding Preserve fixed. Stage~I trains Preserve from multiple
questions per slide; Stage~II trains Focus and MoV to retrieve and
contextualize question-specific evidence against the resulting Bank.

\textbf{Stage I: learning the initial slide impression.}
The CONCH encoders and Qwen parameters remain frozen.
The feature adapter, morphology router, expert transformations, null banks,
expert identities, visual boundary embeddings, and shared projector are
trained with
\begin{equation}
\mathcal{L}_{1}
=
\mathcal{L}_{\mathrm{CQ}}
+
\lambda_p\mathcal{L}_{\mathrm{proto}}
+
\lambda_d\mathcal{L}_{\mathrm{div}} .
\label{eq:stage1_objective}
\end{equation}
$\mathcal{L}_{\mathrm{CQ}}$ supervises answer generation across questions,
$\mathcal{L}_{\mathrm{proto}}$ anchors the semantic experts to prompt-derived
targets, and $\mathcal{L}_{\mathrm{div}}$ discourages excessive overlap in
selection distributions involving free experts.
For each sampled training WSI, the Bank supplies all associated question
micro-batches; their answer gradients are averaged before the parameter
update. Qwen's parameters are frozen, but gradients propagate through its
computations to the visual prefix, training the projector and Preserve.

\textbf{Stage II: learning question-directed focus and fusion.}
Stage~II loads and freezes the complete Preserve path, including the
feature adapter. Hard expert-wise Top-$K$ constructs the Morphology Bank
from the current WSI.
Focus, MoV Fusion, the shared projector, and rank-32 Qwen LoRA parameters
\citep{hulora} are optimized for question-specific answer generation.
The fused prefix provides the primary loss $\mathcal{L}_{\mathrm{fused}}$.
An auxiliary Focus-only view provides direct answer supervision to the
evidence selector: it sets $H_{s,q}=E_{s,q}$ and passes the $K_q$ evidence
tokens through the projector to Qwen, bypassing Bank cross-attention,
evidence self-attention, and the MoV feed-forward sublayers.
With $\zeta\sim\operatorname{Bernoulli}(p_{\mathrm{aux}})$, the objective is
\begin{equation}
\mathcal{L}_{2}
=
\mathcal{L}_{\mathrm{fused}}
+
\zeta\lambda_f\mathcal{L}_{\mathrm{focus}} .
\label{eq:stage2_objective}
\end{equation}
Sampling the auxiliary view limits the number of additional decoder
evaluations during training.
The Qwen base parameters remain frozen; LoRA is applied to the query, key,
value, and output attention projections.

\begin{table*}[t]
\centering
\tablefont
\setlength{\tabcolsep}{1.45pt}
\renewcommand{\arraystretch}{1.03}
\caption{
WSI-Bench VQA results on the official test split.
LLM Visual Tokens reports the number of visual tokens supplied to the
language model where available.
Avg-9 is the unweighted mean of WSI-Precision, WSI-Relevance, and accuracy
across the three task groups.
\ranklegend
}
\label{tab:wsi_bench_vqa}
\adjustbox{max width=\textwidth}{%
\begin{tabular}{lccccccccccc}
\toprule
Method
& \shortstack{LLM Visual\\Tokens $\downarrow$}
& \multicolumn{3}{c}{Morphological Analysis}
& \multicolumn{3}{c}{Diagnosis}
& \multicolumn{3}{c}{Treatment Planning}
& Avg-9 \\
\cmidrule(lr){3-5}
\cmidrule(lr){6-8}
\cmidrule(lr){9-11}
&
& \shortstack{WSI-\\Precision}
& \shortstack{WSI-\\Relevance}
& Accuracy
& \shortstack{WSI-\\Precision}
& \shortstack{WSI-\\Relevance}
& Accuracy
& \shortstack{WSI-\\Precision}
& \shortstack{WSI-\\Relevance}
& Accuracy
& \\
\midrule
GPT-4o
& --
& 0.220 & 0.204 & 0.471
& 0.472 & 0.457 & 0.530
& 0.496 & \second{0.841} & 0.875
& 0.507 \\
Quilt-LLaVA~\citep{seyfioglu2024quilt}
& \second{576}
& 0.448 & 0.447 & 0.947
& 0.586 & 0.604 & 0.849
& 0.788 & 0.816 & \best{1.000}
& 0.721 \\
SlideChat~\citep{chen2025slidechat}
& 20,480
& 0.269 & 0.281 & 0.870
& 0.319 & 0.385 & 0.767
& 0.654 & 0.577 & 0.333
& 0.495 \\
WSI-LLaVA~\citep{liang2025wsi}
& \second{576}
& 0.488 & \second{0.610} & \second{0.951}
& \second{0.610} & \second{0.612} & 0.863
& \best{0.810} & \best{0.845} & \best{1.000}
& \second{0.754} \\
WSI-VQA~\citep{chen2024wsi}
& --
& 0.395 & 0.462 & 0.758
& 0.436 & 0.525 & 0.577
& \second{0.791} & 0.595 & 0.771
& 0.590 \\
MLLM-HWSI~\citep{alawode2026mllm}
& 1,941
& 0.528 & 0.577 & \best{0.989}
& 0.569 & 0.598 & \best{0.962}
& 0.709 & 0.785 & \second{0.986}
& 0.745 \\
HistoSelect~\citep{huang2026act}
& 5,000
& \second{0.538} & 0.589 & 0.946
& 0.542 & 0.587 & 0.858
& 0.766 & 0.801 & 0.979
& 0.734 \\
\midrule
\rowcolor{OursRowBlue}
\textbf{\method}
& \best{96}
& \best{0.592}
& \best{0.645}
& 0.945
& \best{0.621}
& \best{0.639}
& \second{0.875}
& \second{0.791}
& 0.815
& \best{1.000}
& \best{0.769} \\
\bottomrule
\end{tabular}%
}
\vspace{-0.8\baselineskip}
\end{table*}

\textbf{Inference.}
Each request constructs a Bank from the current slide's cached CONCH
features using the frozen adapter and hard Preserve selection. Focus scores
all foreground features for the question and selects 96 evidence anchors.
Two MoV blocks update these anchors before projection and Qwen decoding.
Only the current slide and question are needed; multi-question supervision
is confined to training. The implementation rebuilds the Bank for every
request, including repeated questions about the same slide.
Appendices~\ref{app:reproducibility} and~\ref{app:exact_method} provide
optimization settings, module states, prompts, and selection recurrences.

\section{Experiments}
\label{sec:experiments}

\subsection{Experimental Setup}

\textbf{Benchmarks and data partitions.}
We evaluate within-cohort performance and design choices on WSI-Bench
\citep{liang2025wsi}, then test transfer on SlideBench-BCNB without
target-cohort adaptation \citep{chen2025slidechat}.
WSI-Bench contains 9,850 WSIs and 179,569 question--answer pairs across
30 TCGA cancer types, with an official WSI-level split.
Within the official training split, we hold out 15\% of patients for
checkpoint selection by validation Avg-9.
Each patient's slides and their associated questions remain in the same
training or validation partition.
Main comparisons and controlled ablations use the official test split.
Parameter sweeps are reported descriptively after that decision and are not
used for checkpoint selection or for changing the reported default.

\textbf{Metrics and evaluation tracks.}
WSI-Bench VQA covers morphological analysis, diagnosis, and treatment
planning. WSI-Precision (WSI-P) and WSI-Relevance (WSI-R) evaluate
open-ended answers, while accuracy evaluates closed-ended answers.
\emph{Avg-9} is the unweighted mean of these three metrics across the
three task groups.
For SlideBench-BCNB, we report per-task accuracy and question-count-weighted
Overall accuracy.
For WSI-Bench report generation, the same Stage-II checkpoint is evaluated
under a fixed report-generation instruction using BLEU-1/2/3/4, ROUGE-L,
METEOR, WSI-P, and WSI-R.

\textbf{Features and comparison protocols.}
After foreground filtering, we extract non-overlapping $224\times224$
patches at $20\times$ and encode them with frozen CONCH V1
\citep{lu2024visual}.
Published baseline results follow their respective official protocols.
Unless explicitly varied, controlled comparisons share the cached CONCH
features, Qwen backbone, training annotations and schedule, evaluator,
and official WSI-Bench test split, allowing the ablations to isolate design
choices within the CONCH--Qwen pipeline. Appendix~\ref{app:reproducibility}
provides implementation and evaluation details.

\subsection{Main Results}

\textbf{WSI-Bench VQA.}
On WSI-Bench, \method achieves an Avg-9 of 0.769 with a 96-token visual
prefix, compared with 0.754 for WSI-LLaVA, and leads the pathology-oriented
WSI-P and WSI-R measures in morphological analysis and diagnosis
(Table~\ref{tab:wsi_bench_vqa}). The pattern spans open-ended and closed-ended
evaluation, consistent with an interface that retains both question-specific
evidence and the morphology needed to interpret it.

\begin{table*}[t]
\centering
\tablefont
\setlength{\tabcolsep}{1.85pt}
\renewcommand{\arraystretch}{1.03}
\caption{
Zero-shot accuracy on SlideBench-BCNB.
Overall is weighted by the number of questions in each task, as defined in
Appendix~\ref{app:bcnb_aggregation}. General-model rows are local or
reproduced evaluations under the normalization and endpoint protocol in
Appendix~\ref{app:bcnb_provenance}; they are not presented as verbatim model-
paper scores.
\ranklegend
}
\label{tab:slidebench}
\adjustbox{max width=\textwidth}{%
\begin{tabular}{lcccccccc}
\toprule
Method
& \shortstack{Tumor\\Type}
& \shortstack{ER\\Status}
& \shortstack{PR\\Status}
& \shortstack{HER2\\Expression}
& \shortstack{HER2\\Status}
& Grading
& \shortstack{Molecular\\Subtype}
& Overall \\
\midrule
GPT-4o
& 0.568 & 0.500 & 0.494 & 0.231 & 0.611 & 0.385 & 0.268 & 0.438 \\
GPT-5.4
& 0.604 & 0.504 & 0.499 & 0.294 & 0.596 & 0.424 & 0.284 & 0.458 \\
Gemini-3.1-Pro-preview
& 0.585 & 0.502 & 0.496 & 0.266 & 0.606 & 0.409 & 0.276 & 0.449 \\
Qwen3.5-4B
& 0.494 & 0.334 & 0.273 & 0.298 & 0.662 & \second{0.527} & 0.192 & 0.395 \\
Qwen3.5-9B
& 0.538 & 0.364 & 0.302 & 0.254 & 0.696 & \best{0.592} & 0.214 & 0.420 \\
Quilt-LLaVA~\citep{seyfioglu2024quilt}
& 0.335 & 0.494 & 0.441 & 0.141 & 0.377 & 0.350 & 0.252 & 0.341 \\
SlideChat~\citep{chen2025slidechat}
& 0.899 & 0.730 & 0.730 & 0.251 & 0.719 & 0.231 & 0.206 & 0.544 \\
WSI-LLaVA~\citep{liang2025wsi}
& 0.853 & 0.608 & 0.608 & 0.277 & 0.579 & 0.463 & \second{0.292} & 0.527 \\
MedDr~\citep{he2026towards}
& 0.596 & 0.438 & 0.392 & 0.160 & 0.388 & 0.294 & 0.208 & 0.355 \\
TITAN~\citep{ding2025multimodal}
& 0.852 & 0.424 & 0.417 & 0.244 & 0.494 & 0.454 & 0.225 & 0.444 \\
MedGemma-1.5-4B~\citep{sellergren2026medgemma}
& 0.797 & 0.524 & 0.604 & 0.298 & 0.660 & 0.423 & 0.213 & 0.504 \\
Patho-R1 only~\citep{zhang2026patho}
& 0.422 & 0.546 & 0.504 & 0.146 & 0.366 & 0.355 & 0.267 & 0.373 \\
MLLM-HWSI~\citep{alawode2026mllm}
& \best{0.980} & \best{0.793} & \best{0.796} & \second{0.358} & \second{0.726} & 0.323 & 0.273 & \second{0.612} \\
HistoSelect~\citep{huang2026act}
& 0.915 & 0.749 & 0.748 & 0.288 & 0.693 & 0.248 & 0.225 & 0.558 \\
\midrule
\rowcolor{OursRowBlue}
\textbf{\method}
& \second{0.947} & \second{0.791} & \second{0.751} & \best{0.380} & \best{0.763} & 0.477 & \best{0.344} & \best{0.639} \\
\bottomrule
\end{tabular}%
}
\vspace{-0.7\baselineskip}
\end{table*}

\textbf{SlideBench-BCNB.}
Without target-cohort adaptation, \method reaches 0.639
question-count-weighted Overall accuracy and leads on HER2 expression, HER2
status, and molecular subtype (Table~\ref{tab:slidebench}). This result
supports transfer beyond the WSI-Bench question distribution, although the
best-performing model varies by endpoint.

\begin{table*}[t]
\centering
\tablefont
\setlength{\tabcolsep}{2.45pt}
\renewcommand{\arraystretch}{1.04}
\caption{
WSI-Bench report-generation results.
\ranklegend
}
\label{tab:wsi_report}
\adjustbox{max width=\textwidth}{%
\begin{tabular}{lcccccccc}
\toprule
Method
& BLEU-1
& BLEU-2
& BLEU-3
& BLEU-4
& ROUGE-L
& METEOR
& WSI-P
& WSI-R \\
\midrule
GPT-4o
& 0.202 & 0.069 & 0.030 & 0.016 & 0.132 & 0.167 & 0.067 & 0.138 \\
MI-Gen~\citep{chen2024wsicaption}
& 0.403 & 0.306 & 0.248 & 0.209 & 0.446 & 0.407 & 0.310 & 0.377 \\
Hist-Gen~\citep{guo2024histgen}
& 0.406 & 0.307 & 0.248 & 0.208 & 0.448 & 0.416 & 0.300 & 0.397 \\
Quilt-LLaVA~\citep{seyfioglu2024quilt}
& 0.474 & 0.351 & 0.282 & 0.236 & 0.475 & 0.460 & 0.324 & 0.333 \\
MLLM-HWSI~\citep{alawode2026mllm}
& \second{0.556}
& \second{0.426}
& \second{0.348}
& \second{0.287}
& \second{0.551}
& \second{0.513}
& \second{0.447}
& \second{0.479} \\
SlideChat~\citep{chen2025slidechat}
& 0.439 & 0.310 & 0.277 & 0.191 & 0.463 & 0.422 & 0.123 & 0.168 \\
WSI-LLaVA~\citep{liang2025wsi}
& 0.480 & 0.358 & 0.287 & 0.240 & 0.490 & 0.465 & 0.380 & 0.429 \\
HistoSelect~\citep{huang2026act}
& 0.431 & 0.324 & 0.262 & 0.221 & 0.463 & 0.431 & 0.340 & 0.391 \\
\midrule
\rowcolor{OursRowBlue}
\textbf{\method}
& \best{0.607}
& \best{0.435}
& \best{0.364}
& \best{0.313}
& \best{0.563}
& \best{0.541}
& \best{0.525}
& \best{0.552} \\
\bottomrule
\end{tabular}%
}
\vspace{-1.0\baselineskip}
\end{table*}

\textbf{WSI-Bench report generation.}
The same Stage-II checkpoint leads all eight reported report-generation
metrics, including both lexical and pathology-oriented measures
(Table~\ref{tab:wsi_report}). The shared checkpoint transfers the same visual
interface from focused question answering to longer-form slide description,
where the model must organize multiple findings rather than produce a single
label.

\subsection{Controlled Analyses}
\label{sec:ablation}

Table~\ref{tab:ablation_main} tests stream complementarity, current-slide
multi-question Bank supervision, and fusion. Prefixes use 96 tokens unless
noted; Appendix~\ref{app:ablation_analysis} gives variant definitions and
per-metric results.

\textbf{Preserve and Focus complementarity.}
The full model (0.769 Avg-9) outperforms Focus only (0.721), Preserve only
(0.713), and Bank-restricted Focus (0.718); single-stream scoring reaches
0.739.

\textbf{Bank content and fusion.}
All-QA Preserve (0.769) improves on Single-QA Preserve (0.736); random
same-slide and wrong-slide Banks reduce scores to 0.713 and 0.677.
MoV (0.769) outperforms resampling (0.727), 192-token concatenation (0.739),
and uniform gates (0.756).

\begin{figure*}[t]
\centering
\begin{minipage}[t]{0.49\textwidth}
\centering
\includegraphics[width=\linewidth]{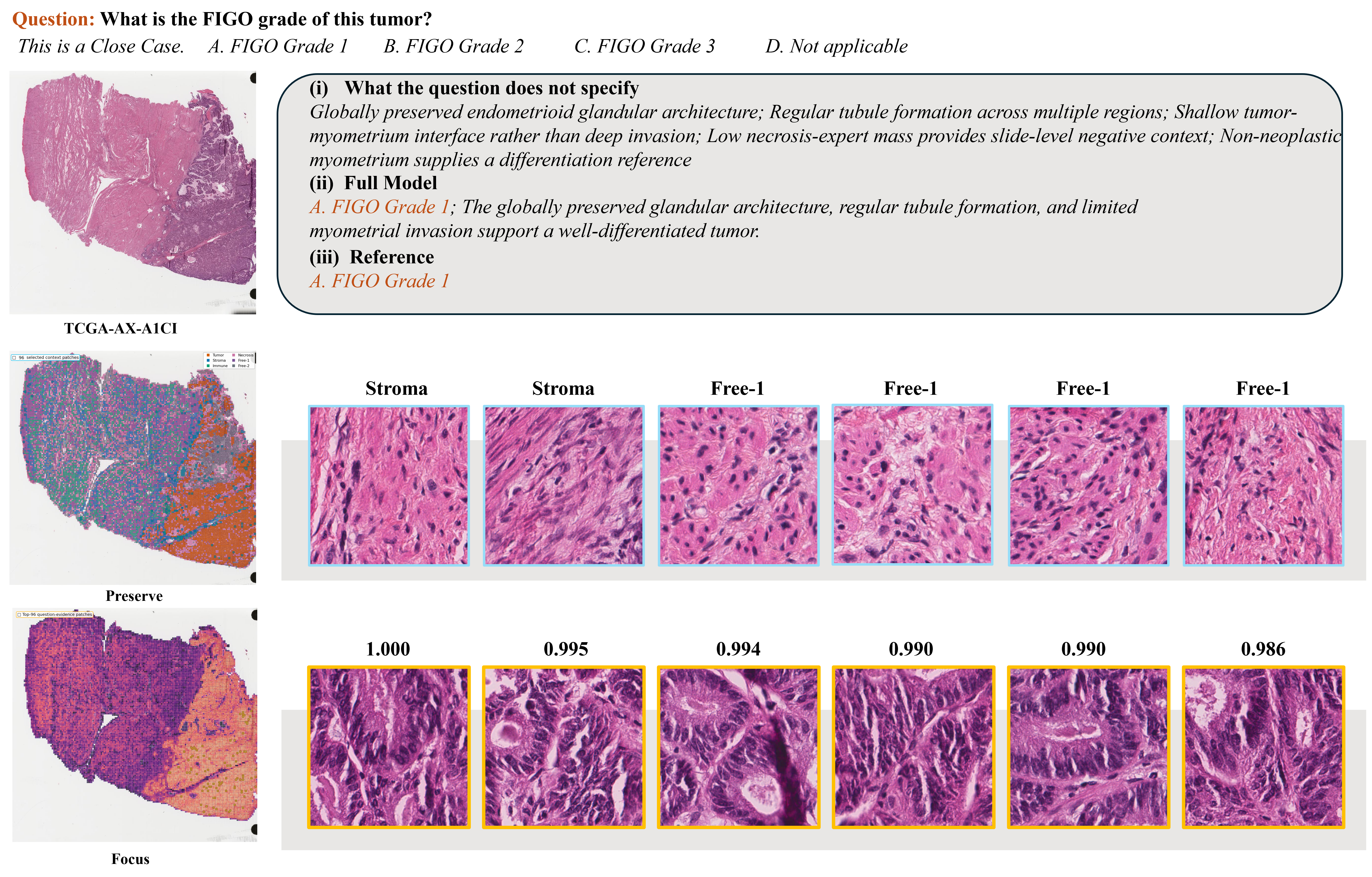}
\captionsetup{font=footnotesize,skip=2pt}
\captionof{figure}{
FIGO grading: Focus selects glands, while Preserve adds tumor and stromal
morphology.
}
\label{fig:case1}
\end{minipage}
\hfill
\begin{minipage}[t]{0.49\textwidth}
\centering
\includegraphics[width=\linewidth]{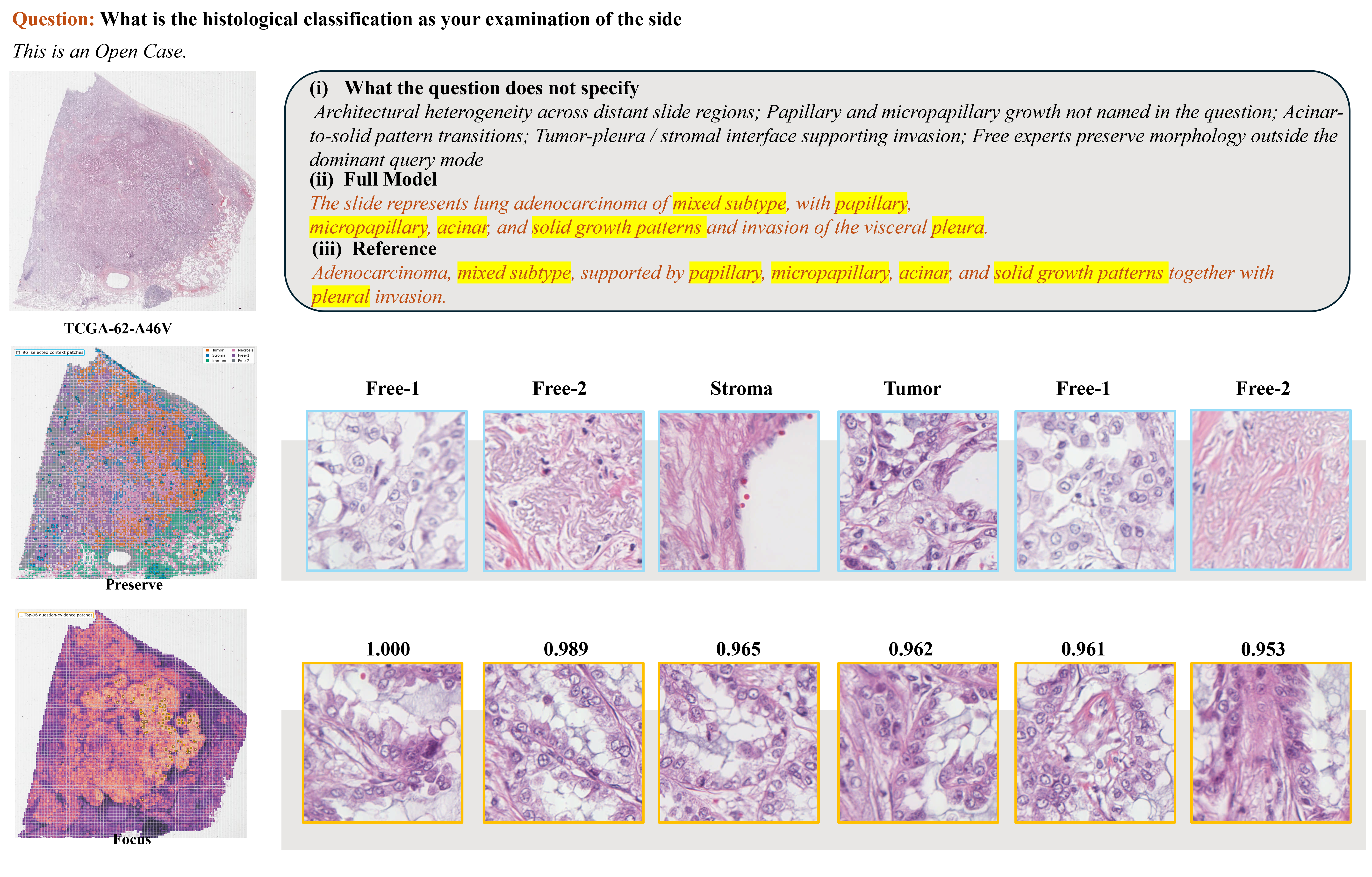}
\captionsetup{font=footnotesize,skip=2pt}
\captionof{figure}{
Lung classification: Focus selects growth patterns, while Preserve adds
heterogeneous morphology.
}
\label{fig:case2}
\end{minipage}
\vspace{-0.6\baselineskip}
\end{figure*}

\begin{table*}[!t]
\centering
\tablefont
\setlength{\tabcolsep}{3.4pt}
\renewcommand{\arraystretch}{1.08}

\caption{
WSI-Bench ablations (Avg-9). LLM prefixes use 96 tokens unless specified.
The configuration tuple denotes (tokens, MoV blocks, semantic/free experts).
Bank-restricted Focus limits retrieval to patches selected by Preserve.
\ranklegend
}
\label{tab:ablation_main}

\begin{tabular*}{\textwidth}{
    @{\extracolsep{\fill}}
    lc@{\qquad}
    lc@{\qquad}
    lc
    @{}
}
\toprule
\multicolumn{2}{c}{\textbf{(a) Preserve \& Focus}}
&
\multicolumn{2}{c}{\textbf{(b) Bank \& Fusion}}
&
\multicolumn{2}{c}{\textbf{(c) Configuration}}
\\
\cmidrule(lr){1-2}
\cmidrule(lr){3-4}
\cmidrule(lr){5-6}
Setting & Avg-9
& Setting & Avg-9
& Setting & Avg-9
\\
\midrule
    Focus only & 0.721
& Random same-slide Bank & 0.713
& Budget $=64$ & 0.760
\\
    Preserve only & 0.713
& Wrong-slide Bank & 0.677
& Budget $=192$ & \best{0.772}
\\
Single-stream & \second{0.739}
& Concat. + Resampler (96) & 0.727
& MoV blocks $=1$ & 0.760
\\
Bank-restricted Focus & 0.718
& Direct concatenation (192) & 0.739
& MoV blocks $=4$ & 0.765
\\
Single-QA Preserve & 0.736
& Uniform expert gate & \second{0.756}
& Experts $=4/0$ & 0.759
\\
\midrule
\rowcolor{OursRowBlue}
\textbf{Full model (all QA)} & \best{0.769}
& \textbf{Full model} & \best{0.769}
& \textbf{Default $(96,2,4/2)$} & \second{0.769}
\\
\bottomrule
\end{tabular*}
\vspace{-0.7\baselineskip}
\end{table*}

\textbf{Budget and model configuration.}
Raising both budgets to 192 yields 0.772; four-block MoV and four
 semantic-only experts score 0.765 and 0.759. Because both selectors scan all
 patches, these are interface rather than retrieval-efficiency comparisons
 (Appendix~\ref{app:efficiency}).

\subsection{Qualitative Analysis}

The two pathologist-reviewed cases illustrate complementary broad review and
question-directed inspection: Focus selects diagnostic patterns, Preserve
supplies broader morphology, and MoV combines both at the evidence positions
(Figures~\ref{fig:case1} and~\ref{fig:case2}). Appendix~\ref{app:qualitative}
compares the full model with either stream and random patches.
% These examples also clarify the mechanism behind the measured gains.
% Focus changes its selected regions with the diagnostic question, whereas
% Preserve contributes slide-level morphology that remains useful across those
% queries. MoV does not merge the paths into a single undifferentiated set;
% it conditions the preserved context on each evidence position, retaining
% the distinction between reusable morphology and question-specific findings.

\section{Conclusion}
\label{sec:conclusion}

We introduced \method, a Preserve--Focus--Fuse framework inspired by
pathologists' broad review, targeted examination, and evidence integration.
Cross-question supervision learns the Morphology Bank, Focus retrieves from
the full foreground set, and MoV fuses both in a 96-token prefix. Results
across WSI-Bench, report generation, and zero-shot SlideBench-BCNB, together
with controlled comparisons, support this design. The resulting principle is
to look beyond explicit cues: retain broad morphology, seek the findings
needed for each question, and bring them together when answering.

\bibliography{references}
\bibliographystyle{plainnat}

\clearpage
\appendix
\setlength{\parskip}{0.3\baselineskip}

\section{Reproducibility and Evaluation Protocol}
\label{app:reproducibility}

This appendix records the implementation and evaluation details omitted from
the main paper. It follows the same order as the model: preserve slide-level
morphology, retrieve question-specific evidence, and fuse the two streams.
Section~\ref{app:exact_method} expands the non-standard selection and
fusion operations, Section~\ref{app:ablation_analysis} reports the
controlled ablations, and Section~\ref{app:efficiency} gives the computational
profiling protocol.

\subsection{Data Partitions and WSI Processing}

\textbf{Data partitions.}
WSI-Bench covers 30 cancer types from TCGA \citep{weinstein2013cancer}.
We follow its official split of 9,642 training WSIs with
175,450 question--answer pairs and 208 test WSIs with 4,119 pairs.
Within the official training split, 15\% of the patients are reserved for
checkpoint selection and the remaining 85\% are used for optimization.
All questions from the same WSI and all WSIs from the same patient remain
in the same training or validation partition.
The partition, parameter initialization, and patch/question sampling use
random seed 42.

The official test split is used for the main benchmark comparisons and
controlled ablations. Parameter-sensitivity results are reported
descriptively and are not used for checkpoint selection.
No test WSI or test question--answer pair is used to update model
parameters.
For report generation, we preserve the WSI-level split and replace the
question with a fixed report-generation instruction.
The WSI-Bench-trained VQA model is evaluated directly on
SlideBench-BCNB without target-cohort training or checkpoint selection.

\subsection{BCNB Score Aggregation}
\label{app:bcnb_aggregation}

The seven-task version of SlideBench-BCNB lists 926 grading questions
and 1,058 questions for each of the other six tasks
\citep{chen2025slidechat}. These unequal task sizes require weighting
task accuracies by question count when reporting Overall accuracy.

For task order (tumor type, ER, PR, HER2 expression, HER2 status,
grading, molecular subtype), the question counts are
$(n_1,\ldots,n_7)=(1058,1058,1058,1058,1058,926,1058)$.

Let $c_t$ be the number of correct answers and $A_t=c_t/n_t$.
Overall is
\begin{equation}
\mathrm{Overall}
=
\frac{\sum_{t=1}^{7}c_t}{7274}
=
\frac{
1058\!\sum_{t\ne\mathrm{grading}} A_t
+
926 A_{\mathrm{grading}}
}{7274}.
\label{eq:bcnb_overall}
\end{equation}

HER2 expression and HER2 status are separate tasks and each receives
weight $1058/7274$.
Overall is computed before rounding task scores.

\subsection{Baseline Provenance and BCNB Evaluation Join}
\label{app:bcnb_provenance}

The general-model rows in Table~\ref{tab:slidebench} use one slide-thumbnail
input per question. GPT-4o, GPT-5.4, and Gemini-3.1-Pro-preview are queried
through their corresponding named endpoints; Qwen3.5-4B and Qwen3.5-9B use the
local checkpoints with the same thumbnail input. Outputs are lower-cased,
stripped of surrounding punctuation, and mapped to the official seven-label
vocabulary; ambiguous or unmappable outputs are counted as incorrect.
These rows are therefore local or reproduced evaluations under the present
answer-normalization and letter-exact BCNB join, not claims that the model
papers reported these exact values.

The resulting compact scores agree, after rounding, with the corresponding
single-thumbnail comparison reported by PathNavigate~\citep{yang2026pathnavigate}.
That comparison reports a merged HER2 score obtained by averaging HER2 status
and HER2 expression. We retain the two component tasks separately in
Table~\ref{tab:slidebench}; each component is computed from its own question
set, and the Overall score is computed from all seven task-level counts before
rounding. The merged value is not used to reconstruct either component.

\begin{table}[H]
\centering
\tablefont
\setlength{\tabcolsep}{2.5pt}
\caption{Use of each benchmark and evaluation track.}
\label{tab:dataset_usage}
\adjustbox{max width=\columnwidth}{%
\begin{tabular}{lccc}
\toprule
Dataset / track
& Optimization
& Model selection
& Evaluation \\
\midrule
WSI-Bench training split
& Stage~I and Stage~II
& 15\% patient hold-out
& Unified multitask training \\

WSI-Bench VQA test
& No
& No
& Morphology, diagnosis, and treatment \\

WSI-Bench report test
& No
& No
& Report generation with the same checkpoint \\

SlideBench-BCNB
& No
& No
& Zero-shot cross-cohort transfer \\
\bottomrule
\end{tabular}%
}
\end{table}

\textbf{Tissue processing and feature extraction.}
WSIs are read at $20\times$ magnification.
Following the CLAM-style tissue-mask procedure
\citep{lu2021data}, the 8-bit HSV saturation channel uses
threshold 8, followed by a median filter of size 7 and morphological closing
with size 4; Otsu thresholding is disabled.
An extracted patch is retained when more than 10\% of its area overlaps the
tissue mask.
Non-overlapping $224\times224$ patches are extracted from retained tissue,
and background tiles are discarded.
Each patch stores its slide identifier, extraction level, integer index,
and level-0 coordinate.
Coordinates are retained for patch recovery and visualization.

Each retained patch is encoded once by the frozen CONCH V1 image encoder.
We cache its 512-dimensional pre-normalization feature and its
$L_2$-normalized copy.
The former is adapted for routing and fusion; the latter remains in the
CONCH contrastive space for morphology anchoring and question--patch
matching.
All controlled experiments use the same cached patch features, so the
ablations change evidence organization while holding the visual input fixed.

\subsection{Architecture and Optimization}

\textbf{Default architecture.}
All trainable visual modules use a 512-dimensional workspace.
Preserve contains four semantic and two free experts, each retaining 16 patch
anchors. Focus selects 96 evidence anchors, and MoV uses two blocks with
eight attention heads to fuse the views.
The expert FFN has dimensions $512\rightarrow1024\rightarrow512$, and the
visual projector has dimensions $512\rightarrow2048\rightarrow3584$.
Qwen2.5-7B-Instruct is used through a continuous visual prefix.
Its base parameters remain frozen, and Stage~II adds rank-32 LoRA to the
query, key, value, and output projections of each self-attention layer.
The same CONCH encoder is used in every controlled comparison, separating
interface changes from backbone changes.

\begin{table}[t]
\centering
\tablefont
\setlength{\tabcolsep}{4pt}
\renewcommand{\arraystretch}{1.05}
\caption{Module states during training and inference.}
\label{tab:app_module_states}
\begin{tabular}{lccc}
\toprule
Component & Stage~I & Stage~II & Inference \\
\midrule
CONCH image encoder & Frozen & Frozen & Frozen \\
CONCH text encoder & Frozen & Frozen & Frozen \\
Feature adapter & Train & Frozen & Frozen \\
Morphology router & Train & Frozen & Frozen \\
Expert FFNs and identities & Train & Frozen & Frozen \\
Semantic null banks & Train & Frozen & Frozen \\
Preserve selection & ST Top-$K$ & Hard Top-$K$ & Hard Top-$K$ \\
Focus scorer and selector & Off & Train & Frozen \\
MoV Fusion & Off & Train & Frozen \\
Visual projector & Train & Train & Frozen \\
Qwen base parameters & Frozen & Frozen & Frozen \\
Qwen LoRA & Off & Train & Frozen \\
\bottomrule
\end{tabular}
\end{table}

\textbf{Optimization.}
Both stages use AdamW, bfloat16 mixed precision, gradient clipping at
$1.0$, a 5\% linear warm-up, cosine learning-rate decay, and weight decay
$0.05$.
Stage~I runs for five epochs.
The adapter, router, and expert modules use learning rate
$1\times10^{-4}$, while the projector uses $5\times10^{-5}$.
One WSI is sampled per GPU, and gradients from its question micro-batches
are averaged before each Preserve update.
Stage~II runs for three epochs with learning rate $5\times10^{-5}$ for
Focus, MoV, and the projector and $2\times10^{-5}$ for Qwen LoRA.
Gradient accumulation gives an effective batch size of 32
question--answer pairs.
The checkpoint with the highest validation Avg-9 is used for evaluation.

We set $\lambda_p=0.01$, $\lambda_d=0.005$, and $\lambda_f=0.3$.
The Focus-only auxiliary branch is sampled with probability $0.5$.
The Preserve relaxation temperature is annealed from $1.0$ to $0.1$
during Stage~I.  The router temperature is $\tau_r=1.0$, the semantic-target
temperature is $\tau_p=0.07$, and the presence temperature is
$\tau_\rho=0.05$.  The default Focus temperature is $\tau_q=0.07$, and
dropout is $0.1$ in the adapter, expert FFNs, and MoV blocks.
Training uses three 40-GB NVIDIA A100 GPUs.
AdamW uses $(\beta_1,\beta_2)=(0.9,0.999)$ and optimizer
$\epsilon=10^{-8}$; the Stage-I selection temperature decreases linearly
from 1.0 to 0.1 over optimizer updates.

\textbf{Decoding and evaluation.}
Decoding and scoring follow the WSI-Bench and SlideBench-BCNB protocols.
VQA uses greedy decoding with at most 128 new tokens; report generation
uses at most 512.
Closed-ended answers are lower-cased, stripped of surrounding punctuation,
and mapped to the official label vocabulary; an output that cannot be mapped
unambiguously to one allowed label is counted as incorrect.
WSI-P and WSI-R use the GPT-4o claim-scoring prompts released with WSI-LLaVA
and deterministic decoding with temperature 0 and top-$p=1$.
Report generation uses greedy decoding with the fixed instruction
``Describe the histopathological findings in this whole-slide image.''
BLEU-1/2/3/4, ROUGE-L, and METEOR use the official WSI-Bench scripts;
BLEU follows the original n-gram precision-based definition
\citep{papineni2002bleu}.

\subsection{Morphology Prompt Ensembles}

Each semantic expert uses a short prompt ensemble to define its morphology
target in the frozen CONCH image--text space.
The normalized CONCH text embeddings in each row of
Table~\ref{tab:app_prompts} are averaged and re-normalized.
The two free experts do not use text anchors.

\begin{table*}[t]
\centering
\tablefont
\setlength{\tabcolsep}{4pt}
\renewcommand{\arraystretch}{1.06}
\caption{Prompt ensembles for the four semantic morphology experts.}
\label{tab:app_prompts}
\begin{tabular}{lp{0.78\textwidth}}
\toprule
Expert & Prompt ensemble \\
\midrule
Tumor &
\texttt{a histopathology image of tumor tissue};
\texttt{malignant neoplastic cells};
\texttt{invasive tumor morphology};
\texttt{a tumor-rich histology region} \\
Stroma &
\texttt{a histopathology image of stromal tissue};
\texttt{fibrous connective tissue stroma};
\texttt{desmoplastic tumor-associated stroma};
\texttt{a stroma-rich histology region} \\
Immune infiltration &
\texttt{a histopathology image with immune-cell infiltration};
\texttt{lymphocyte-rich tissue};
\texttt{tumor-infiltrating lymphocytes};
\texttt{an inflammatory cell-rich histology region} \\
Necrosis &
\texttt{a histopathology image of necrotic tissue};
\texttt{tumor necrosis};
\texttt{coagulative necrosis and cellular debris};
\texttt{a necrosis-rich histology region} \\
\bottomrule
\end{tabular}
\end{table*}

\section{Exact Method Details}
\label{app:exact_method}

\subsection{Notation}

The notation below fixes the dimensions of the visual interface and keeps the
two evidence roles explicit: cross-question answer losses train Preserve to
select patches that form the slide-level initial impression, whereas Focus
retrieves evidence for the current question. MoV uses the resulting Bank to
contextualize that evidence, as described in the selection and fusion
operations below.

\begin{table}[t]
\centering
\tablefont
\setlength{\tabcolsep}{4pt}
\renewcommand{\arraystretch}{1.05}
\caption{Principal notation and tensor shapes.}
\label{tab:app_notation}
\begin{tabular}{lll}
\toprule
Symbol & Meaning & Shape \\
\midrule
$N_s$ & foreground patches in slide $s$ & scalar \\
$X_s$ & adapted content features & $N_s\times512$ \\
$Z_s^{\mathrm{vl}}$ & normalized CONCH features & $N_s\times512$ \\
$q_c$ & normalized question embedding & $512$ \\
$B_{s,e}$ & bank of expert $e$ & $k_e\times512$ \\
$M_s$ & concatenated Morphology Bank & $K_m\times512$ \\
$E_{s,q}$ & Focus-selected evidence & $K_q\times512$ \\
$H_{s,q}$ & MoV output & $K_q\times512$ \\
$\rho_{s,e}$ & morphology presence & scalar \\
\bottomrule
\end{tabular}%
\end{table}

\subsection{Hard-Forward, Relaxed-Backward Top-\texorpdfstring{$K$}{K}}
\label{app:relaxed_topk}

Preserve and Focus use the same subset operator.
For scores $\mathbf a\in\mathbb R^N$ and budget $K$,
$\mathbf P^{\mathrm{hard}}(\mathbf a,K)\in\{0,1\}^{K\times N}$
contains one ordered one-hot row for each of the $K$ highest-scoring
indices.
The relaxed path starts from
$\boldsymbol\ell^{(1)}=\mathbf a$ and constructs one soft selection row
per slot:
\begin{equation}
\mathbf p^{(r)}
=
\operatorname{Softmax}
\left(
\boldsymbol\ell^{(r)}/\tau
\right).
\label{eq:app_relaxed_slot}
\end{equation}
Already selected candidates are suppressed through
\begin{equation}
\boldsymbol\ell^{(r+1)}
=
\boldsymbol\ell^{(r)}
+
\log\!\left(
1-\mathbf p^{(r)}+\epsilon
\right),
\qquad \epsilon=10^{-6}.
\label{eq:app_relaxed_suppression}
\end{equation}
Stacking the rows gives
$\mathbf P^{\mathrm{soft}}(\mathbf a,K)$, and the straight-through matrix
is
\begin{equation}
\mathbf P^{\mathrm{ST}}
=
\mathbf P^{\mathrm{soft}}
+
\operatorname{stopgrad}
\left(
\mathbf P^{\mathrm{hard}}-\mathbf P^{\mathrm{soft}}
\right).
\label{eq:app_st_matrix}
\end{equation}
The forward value is the hard subset, while gradients follow the relaxed
matrix.
Before each Softmax, the largest residual logit is subtracted, and
$1-\mathbf p^{(r)}$ is clamped below by $\epsilon$.
The temperature is the linearly annealed Stage-I Preserve temperature when
constructing $M_s$ and is $\tau_q=0.07$ for the Stage-II Focus selector.
Only hard selection is evaluated after Preserve is frozen and at test time.
The operator above specifies the fixed-budget, hard-forward selection used in
our experiments.

\subsection{Semantic Presence and Expert Diversity}

For semantic expert $e$, let $p_e$ be its normalized prompt-ensemble
embedding.
Patch--anchor similarities
$a_{s,i,e}^{\mathrm{sem}}=z_{s,i}^{\top}p_e$ define
\begin{equation}
t_{s,i,e}
=
\frac{\exp(a_{s,i,e}^{\mathrm{sem}}/\tau_p)}
{\sum_{i'=1}^{N_s}\exp(a_{s,i',e}^{\mathrm{sem}}/\tau_p)}.
\label{eq:app_semantic_target}
\end{equation}
The mean of the 16 largest similarities is converted to a soft presence
value,
where $I^p_{s,e}$ contains those 16 indices and
$\bar a^{\mathrm{sem}}_{s,e}=16^{-1}\sum_{i\in I^p_{s,e}}
a^{\mathrm{sem}}_{s,i,e}$,
\begin{equation}
\rho_{s,e}
=
\sigma\!\left(
\frac{\bar a_{s,e}^{\mathrm{sem}}-\delta_e}{\tau_\rho}
\right).
\label{eq:app_presence}
\end{equation}
For each semantic expert, candidate values for $\delta_e$ are the 25th, 50th,
and 75th percentiles of the training-slide distribution of mean top-16
prompt similarities.  The validation split selects the final value; in the
reported configuration the 50th percentile is selected for all four semantic
experts, and the resulting four raw thresholds are stored with the checkpoint.
Presence weights the prototype loss and the interpolation with the
semantic expert's null bank.
Free experts use $\rho_{s,e}=1$.

For the semantic experts, the router is trained against the normalized
prompt-similarity distribution with the presence-weighted objective
\begin{equation}
\mathcal L_{\mathrm{proto}}^{(s)}:=
\frac{\sum_{e\in\mathcal E_{\mathrm{sem}}}\rho_{s,e}
D_{\mathrm{KL}}(t_{s,:,e}\Vert g_{s,:,e})}
{\sum_{e\in\mathcal E_{\mathrm{sem}}}\rho_{s,e}+\epsilon}.
\label{eq:app_proto_loss}
\end{equation}

Let $\boldsymbol\mu_{s,e}$ be the mean of the 16 relaxed selection rows
for expert $e$.
The diversity term is
\begin{equation}
\mathcal L_{\mathrm{div}}^{(s)}
=
\frac{1}{|\mathcal C|}
\sum_{(e,e')\in\mathcal C}
\left[
\max\!\left(
0,
\operatorname{cos}
(\boldsymbol\mu_{s,e},\boldsymbol\mu_{s,e'})
-\delta_{\mathrm{ov}}
\right)
\right]^2,
\label{eq:app_diversity}
\end{equation}
where $\mathcal C$ contains the free--free pair and each
semantic--free pair.
We use $\delta_{\mathrm{ov}}=0.5$.
Semantic--semantic pairs are excluded because overlap at composite tissue
regions is meaningful.
When $\mathcal C=\varnothing$, we define
$\mathcal L_{\mathrm{div}}^{(s)}=0$ rather than evaluating the empty average.

\subsection{MoV and Visual-Prefix Implementation}

Our implementation projects the fused evidence into Qwen's token-embedding
space as follows.
The visual projector is a two-layer MLP with dimensions
$512\rightarrow2048\rightarrow3584$:
\begin{equation}
\Pi(H)=W_{p,2}\operatorname{GELU}\!\left(W_{p,1}\operatorname{LN}(H)+b_{p,1}\right)+b_{p,2}.
\label{eq:visual_projector}
\end{equation}
It maps each 512-dimensional visual token to the 3584-dimensional Qwen
hidden space.
Projected tokens are inserted between learned
\texttt{<vis\_start>} and \texttt{<vis\_end>} embeddings before the
question tokens.
We use Qwen's standard token-embedding interface with one-dimensional
positions and causal attention, and apply the loss only to answer tokens.

Each MoV block contains six expert-specific cross-attention modules.
The evidence sequence provides queries, and one 16-token expert bank
provides keys and values.
For block $\ell$, expert $e$ computes
$A_{s,q,e}^{(\ell)}=\operatorname{MHA}_{e}^{(\ell)}
(\operatorname{LN}(H_{s,q}^{(\ell)}),
\operatorname{LN}(B_{s,e}),\operatorname{LN}(B_{s,e}))$.
For evidence position $k$, the gate input for expert $e$ is
$\xi_{s,q,k,e}=[\operatorname{LN}(H_{s,q,k});q_c;\bar B_{s,e};
\eta_e;\rho_{s,e}]$. Its logit is
$g_{s,q,k,e}=w_g^\top\operatorname{GELU}(W_g\xi_{s,q,k,e}+b_g)+b_{g,e}$;
a separate GELU MLP maps $[\operatorname{LN}(H_{s,q,k});q_c]$ to the
null logit $g_{s,q,k,0}$.
Writing these logits as $g_{s,q,k,e}$ and $g_{s,q,k,0}$, respectively, the
gate is
\begin{equation}
\alpha_{s,q,k,:}=\operatorname{Softmax}
\left(g_{s,q,k,0},g_{s,q,k,1},\ldots,g_{s,q,k,6}\right),
\qquad \sum_{e=0}^{6}\alpha_{s,q,k,e}=1.
\label{eq:mov_gate}
\end{equation}
The null path has zero attention output and therefore lets an evidence token
retain its current state.
The gated update is followed by pre-normalized evidence self-attention and
a residual SwiGLU block.
Only evidence positions are propagated between blocks, so the output
length remains $K_q$.
MoV retains the Focus-selected positions and updates their content; the
resampling and concatenation controls below test whether this positional
conditioning is necessary.

\subsection{Training and Inference Data Flow}

In Stage~I, one Morphology Bank is computed from each sampled WSI and shared by
all of its question micro-batches.
Cross-question answer losses, prototype alignment, and expert diversity
update the adapter, router, expert transformations, null banks, identities,
and projector while CONCH and Qwen remain frozen.

In Stage~II, the complete Preserve path is frozen.
Hard expert-wise Top-$K$ instantiates one Bank for each current WSI.
Focus scores all retained foreground patches for each question, and
ST Top-$K$ retrieves 96 Focus evidence anchors.
MoV reads only the Bank produced from the same WSI.
The fused and Focus-only visual prefixes train Focus, MoV, the projector,
and Qwen LoRA.

At inference, the frozen Preserve network constructs a new Bank for every
request, including subsequent questions about the same slide; no Bank is
cached across questions. Each question then invokes full-set Focus scoring,
hard Top-$K$, two MoV blocks, and Qwen decoding from the resulting 96-token
prefix.

\section{Ablation Analysis}
\label{app:ablation_analysis}

\subsection{Complete Ablation Breakdown}
\label{app:ablation_full}

These ablations distinguish the operations in the Preserve--Focus--Fuse
paradigm: how Preserve selects morphology, how Focus selects
question-specific evidence, and how Bank information reaches the language
model.
The controls test evidence organization within WSI-VQA while holding the
backbone and evaluation split fixed. Prefixes contain 96 tokens except for direct
concatenation, which uses 192.

\textbf{Operational definitions and training status.}
All ablations use the same foreground-patch set, CONCH features, optimizer
schedule, five Stage-I epochs, three Stage-II epochs, and validation protocol.
Variants that change Preserve, its candidate set, or its
supervision are
retrained through both stages.  Fusion-only variants reuse the same frozen
Stage-I checkpoint and retrain Stage~II.  Random same-slide Bank, Wrong-slide
Bank, and Random Focus are inference-time interventions on the full checkpoint
and introduce no fitted parameters.

\emph{Focus only} passes the 96 Focus-selected evidence tokens directly to
the projector, without a Morphology Bank or MoV update. \emph{Preserve only}
passes the 96-token Morphology Bank through the projector without
question-conditioned Focus retrieval. \emph{Single-stream scoring} replaces
the parallel Preserve--Focus paths with one question-conditioned learned
selector over all retained foreground patches and emits a single 96-token
stream. \emph{Bank-restricted Focus} limits retrieval to patch locations
selected by Preserve. Compared with full-slide Focus, this control evaluates
the effect of restricting question-directed inspection to the initially
preserved regions.

For \emph{Random Focus}, 96 foreground-patch indices are sampled uniformly
without replacement once per slide and held fixed across that slide's
questions.  For \emph{Random same-slide Bank}, each expert's selected indices
are replaced by uniformly sampled indices from the same slide before the
existing expert transformations and fusion.  \emph{Wrong-slide Bank} uses a
fixed random permutation of the test slides with no self-matches and applies
the donor slide's Bank before the unchanged MoV and language-model stages.

\emph{Global pooled Bank + residual} averages all current-slide Bank tokens,
broadcasts the pooled vector to the original $K_m$ positions, and adds a
learned $512$-dimensional residual adapter.  The residual adapter is trained
in Stage~II while Preserve remains frozen.  \emph{Concat. + Resampler (96)}
concatenates $M_s$ and $E_{s,q}$ into 192 tokens, then applies a two-block
eight-head latent resampler with 96 learned queries, model width 512, and
FFN width $512\!\rightarrow\!2048\!\rightarrow\!512$ to return 96 tokens.
The resampler and downstream projector are trained in Stage~II.  \emph{Direct
concatenation (192)} supplies the same 192 concatenated tokens directly to the
language-side projector without resampling.

\emph{Uniform expert gate} retains the null path.  If $\alpha_0$ is the
learned null-path weight at an evidence position, each of the six real experts
receives $(1-\alpha_0)/6$ and the null path keeps $\alpha_0$.  \emph{Shared
expert FFN} ties the six expert-transformation weights while retaining
expert-specific routing and quotas; this architecture variant is retrained
through both stages.

For the Single-QA, 2-QA, 4-QA, and All-QA rows, questions are first placed in
a fixed random order for each WSI.  The smaller settings use nested prefixes of
that order, while All-QA uses every available question.  Stage~I still makes
five passes over the same WSI set, and the within-slide question losses are
averaged before each Preserve update.  Thus the comparison changes the number
of distinct QA answers supervising one slide while keeping the number of
slide-level optimization epochs and the answer-loss normalization fixed.

\begin{table*}[t]
\centering
\tablefont
\setlength{\tabcolsep}{1.25pt}
\renewcommand{\arraystretch}{1.02}
\caption{
Complete ablation and intervention results on WSI-Bench.
Direct concatenation uses 192 language-side visual tokens; all other
variants use 96. \ranklegend
}
\label{tab:app_ablation_full}
\adjustbox{max width=\textwidth}{%
\begin{tabular}{lcccccccccc}
\toprule
Variant
& \multicolumn{3}{c}{Morphological Analysis}
& \multicolumn{3}{c}{Diagnosis}
& \multicolumn{3}{c}{Treatment Planning}
& Avg-9 \\
\cmidrule(lr){2-4}\cmidrule(lr){5-7}\cmidrule(lr){8-10}
& WSI-P & WSI-R & Acc.
& WSI-P & WSI-R & Acc.
& WSI-P & WSI-R & Acc.
& \\
\midrule
Focus only
& 0.520 & 0.574 & 0.919 & 0.583 & 0.599 & 0.842
& 0.732 & 0.762 & 0.958 & 0.721 \\
Preserve only
& 0.533 & 0.581 & 0.916 & 0.543 & 0.561 & 0.820
& 0.741 & 0.764 & 0.958 & 0.713 \\
Random same-slide Bank
& 0.510 & 0.570 & 0.908 & 0.574 & 0.583 & 0.835
& 0.725 & 0.754 & 0.958 & 0.713 \\
Random Focus
& 0.505 & 0.553 & 0.904 & 0.497 & 0.530 & 0.803
& 0.698 & 0.731 & 0.917 & 0.682 \\
Single-stream scoring
& 0.553 & 0.613 & 0.922 & 0.586 & 0.605 & 0.851
& 0.759 & 0.783 & \second{0.979} & 0.739 \\
Bank-restricted Focus
& 0.525 & 0.582 & 0.908 & 0.573 & 0.591 & 0.839
& 0.731 & 0.755 & 0.958 & 0.718 \\
\midrule
Single-QA Preserve
& 0.545 & 0.595 & 0.924 & 0.591 & 0.610 & 0.855
& 0.752 & 0.773 & \second{0.979} & 0.736 \\
2 QA per WSI
& 0.564 & 0.612 & 0.931 & 0.601 & 0.619 & 0.861
& 0.762 & 0.791 & \best{1.000} & 0.749 \\
4 QA per WSI
& 0.575 & 0.626 & \second{0.938} & \second{0.610} & \second{0.629} & \second{0.867}
& \second{0.777} & \second{0.800} & \best{1.000} & \second{0.758} \\
All QA per WSI
& \best{0.592} & \best{0.645} & \best{0.945} & \best{0.621} & \best{0.639} & \best{0.875}
& \best{0.791} & \best{0.815} & \best{1.000} & \best{0.769} \\
\midrule
Wrong-slide Bank
& 0.476 & 0.533 & 0.890 & 0.526 & 0.544 & 0.808
& 0.689 & 0.710 & 0.917 & 0.677 \\
Global pooled Bank + residual
& 0.533 & 0.578 & 0.912 & 0.578 & 0.592 & 0.840
& 0.725 & 0.755 & 0.958 & 0.719 \\
Concatenate + Resampler (96)
& 0.548 & 0.597 & 0.918 & 0.580 & 0.592 & 0.845
& 0.739 & 0.766 & 0.958 & 0.727 \\
Direct concatenation (192)
& 0.557 & 0.611 & 0.926 & 0.590 & 0.609 & 0.852
& 0.748 & 0.779 & \second{0.979} & 0.739 \\
Uniform expert gate
& \second{0.580} & \second{0.632} & 0.936 & 0.607 & 0.625 & 0.864
& 0.770 & 0.790 & \best{1.000} & 0.756 \\
Shared expert FFN
& 0.569 & 0.622 & 0.930 & 0.596 & 0.613 & 0.857
& 0.761 & 0.784 & \best{1.000} & 0.748 \\
\midrule
\rowcolor{OursRowBlue}
\textbf{Full model}
& \best{0.592} & \best{0.645} & \best{0.945}
& \best{0.621} & \best{0.639} & \best{0.875}
& \best{0.791} & \best{0.815} & \best{1.000}
& \best{0.769} \\
\bottomrule
\end{tabular}%
}
\end{table*}

\subsection{Parameter Sensitivity}
\label{app:parameter_sensitivity}

\begin{table*}[t]
\centering
\tablefont
\setlength{\tabcolsep}{1.25pt}
\renewcommand{\arraystretch}{1.02}
\caption{
Descriptive parameter sensitivity on the WSI-Bench test split.
Bold settings indicate the default configuration; these sweeps are reported
for characterization and are not used for checkpoint selection. Within each
parameter sweep, the best Avg-9 is in \textcolor{BestRed}{red} and the second
best is in \textcolor{SecondBlue}{blue}.
}
\label{tab:app_parameter_compact}
\adjustbox{max width=\textwidth}{%
\begin{tabular}{llcccccccccc}
\toprule
Parameter & Setting
& \multicolumn{3}{c}{Morphological Analysis}
& \multicolumn{3}{c}{Diagnosis}
& \multicolumn{3}{c}{Treatment Planning}
& Avg-9 \\
\cmidrule(lr){3-5}\cmidrule(lr){6-8}\cmidrule(lr){9-11}
&
& WSI-P & WSI-R & Acc.
& WSI-P & WSI-R & Acc.
& WSI-P & WSI-R & Acc.
& \\
\midrule
$(K_q=K_m)$ & 32
& 0.560 & 0.613 & 0.932 & 0.591 & 0.610 & 0.856 & 0.765 & 0.793 & 0.958 & 0.742 \\
& 64
& 0.583 & 0.635 & 0.941 & 0.612 & 0.630 & 0.869 & 0.783 & 0.808 & 0.979 & 0.760 \\
& \textbf{96}
& 0.592 & 0.645 & 0.945 & 0.621 & 0.639 & 0.875 & 0.791 & 0.815 & 1.000 & 0.769 \\
& 128
& 0.594 & 0.648 & 0.946 & 0.623 & 0.641 & 0.876 & 0.793 & 0.818 & 1.000 & \second{0.771} \\
& 192
& 0.596 & 0.649 & 0.947 & 0.625 & 0.642 & 0.877 & 0.794 & 0.818 & 1.000 & \best{0.772} \\
\midrule
MoV depth 
& 1
& 0.585 & 0.636 & 0.940 & 0.613 & 0.631 & 0.869 & 0.781 & 0.806 & 0.979 & 0.760 \\
& \textbf{2}
& 0.592 & 0.645 & 0.945 & 0.621 & 0.639 & 0.875 & 0.791 & 0.815 & 1.000 & \best{0.769} \\
& 3
& 0.590 & 0.644 & 0.944 & 0.620 & 0.638 & 0.874 & 0.789 & 0.813 & 1.000 & \second{0.768} \\
& 4
& 0.590 & 0.643 & 0.944 & 0.619 & 0.637 & 0.873 & 0.788 & 0.812 & 0.979 & 0.765 \\
\midrule
Semantic/Free & 2/0 (tumor, stroma)
& 0.565 & 0.616 & 0.932 & 0.603 & 0.620 & 0.860 & 0.767 & 0.790 & 0.979 & 0.748 \\
& 4/0, total 4
& 0.581 & 0.633 & 0.939 & 0.613 & 0.631 & 0.869 & 0.781 & 0.805 & 0.979 & 0.759 \\
& \textbf{4/2, total 6}
& 0.592 & 0.645 & 0.945 & 0.621 & 0.639 & 0.875 & 0.791 & 0.815 & 1.000 & \best{0.769} \\
& 4/4, total 8
& 0.589 & 0.642 & 0.943 & 0.619 & 0.637 & 0.873 & 0.788 & 0.812 & 1.000 & \second{0.767} \\
\midrule
$(\lambda_p)$ & 0
& 0.565 & 0.623 & 0.934 & 0.601 & 0.620 & 0.861 & 0.777 & 0.799 & 0.979 & 0.751 \\
& 0.001
& 0.578 & 0.632 & 0.939 & 0.609 & 0.628 & 0.867 & 0.783 & 0.807 & 0.979 & 0.758 \\
& \textbf{0.01}
& 0.592 & 0.645 & 0.945 & 0.621 & 0.639 & 0.875 & 0.791 & 0.815 & 1.000 & \best{0.769} \\
& 0.05
& 0.586 & 0.639 & 0.942 & 0.616 & 0.634 & 0.872 & 0.787 & 0.812 & 0.979 & \second{0.763} \\
& 0.1
& 0.570 & 0.626 & 0.935 & 0.604 & 0.623 & 0.865 & 0.779 & 0.805 & 0.958 & 0.752 \\
\midrule
Expert access & Top-1
& 0.575 & 0.630 & 0.936 & 0.602 & 0.623 & 0.863 & 0.769 & 0.791 & 0.958 & 0.750 \\
& Top-2
& 0.584 & 0.638 & 0.941 & 0.613 & 0.630 & 0.870 & 0.781 & 0.804 & 0.958 & 0.758 \\
& Top-3
& 0.590 & 0.643 & 0.944 & 0.619 & 0.637 & 0.873 & 0.788 & 0.812 & 0.979 & \second{0.765} \\
& \textbf{All-6}
& 0.592 & 0.645 & 0.945 & 0.621 & 0.639 & 0.875 & 0.791 & 0.815 & 1.000 & \best{0.769} \\
& Uniform
& 0.580 & 0.632 & 0.936 & 0.607 & 0.625 & 0.864 & 0.770 & 0.790 & 1.000 & 0.756 \\
\midrule
$(\tau_q)$ & 0.03
& 0.581 & 0.634 & 0.940 & 0.610 & 0.628 & 0.869 & 0.783 & 0.807 & 0.979 & 0.759 \\
& 0.05
& 0.590 & 0.643 & 0.944 & 0.618 & 0.636 & 0.873 & 0.789 & 0.813 & 0.979 & \second{0.765} \\
& \textbf{0.07}
& 0.592 & 0.645 & 0.945 & 0.621 & 0.639 & 0.875 & 0.791 & 0.815 & 1.000 & \best{0.769} \\
& 0.1
& 0.588 & 0.641 & 0.943 & 0.617 & 0.634 & 0.873 & 0.788 & 0.810 & 1.000 & \second{0.766} \\
& 0.2
& 0.580 & 0.633 & 0.940 & 0.608 & 0.625 & 0.868 & 0.782 & 0.807 & 0.979 & 0.758 \\
\midrule
FFN expansion & 1
& 0.584 & 0.636 & 0.939 & 0.612 & 0.629 & 0.869 & 0.780 & 0.803 & 0.979 & 0.759 \\
& \textbf{2}
& 0.592 & 0.645 & 0.945 & 0.621 & 0.639 & 0.875 & 0.791 & 0.815 & 1.000 & \second{0.769} \\
& 4
& 0.593 & 0.646 & 0.946 & 0.622 & 0.640 & 0.876 & 0.792 & 0.815 & 1.000 & \best{0.770} \\
\bottomrule
\end{tabular}%
}
\end{table*}

The expanded sensitivity table provides a descriptive view of the operating
region around the default configuration. Increasing each stream from 32 to
96 tokens is associated with a 0.027 Avg-9 increase, whereas doubling the
default budget to 192 changes Avg-9 by 0.003. Two MoV blocks and four
semantic plus two free experts correspond to the reported default setting.
The prototype and Focus-temperature sweeps show their highest displayed
values near the default settings, and an FFN expansion of four changes Avg-9
by only 0.001 relative to expansion two.
For the joint budget sweep, $(K_m,K_q)$ changes together. With six experts,
the Bank quotas at budgets 32, 64, 96, 128, and 192 are respectively
$(6,6,5,5,5,5)$, $(11,11,11,11,10,10)$, $(16,16,16,16,16,16)$,
$(22,22,21,21,21,21)$, and $(32,32,32,32,32,32)$, ordered as tumor,
stroma, immune infiltration, necrosis, free 1, and free 2.
For expert-count sweeps, both budgets remain 96, giving equal quotas of
48, 24, 16, or 12 for 2, 4, 6, or 8 experts.
Configurations that change Preserve are retrained through both stages;
Stage-II-only changes reuse the same Stage-I checkpoint.

\section{Efficiency and Scaling}
\label{app:efficiency}

\subsection{Profiling Protocol}

These measurements characterize the visual interface under a fixed profiling
protocol rather than a matched-backbone comparison.
We use pre-extracted CONCH features, batch size one, bfloat16 inference, and
at most 128 generated answer tokens. Offline WSI reading, tissue
segmentation, patch extraction, and CONCH feature extraction are excluded.
For each fixed input and patch count, we discard 20 warm-up requests and
measure 100 requests, synchronizing the GPU before and after each timed
region. The timed region includes visual selection, fusion, language-model
prefill, and generation with EOS stopping and at most 128 new tokens.
Latency is summarized by the median and the empirical 2.5th--97.5th
percentile interval on an NVIDIA A100 40-GB GPU. Each single-question
measurement includes Preserve execution and Bank construction for that
request; no Bank is reused across questions.
The fixed-input cost profile uses 20,480 foreground patches; FLOPs count
matrix multiplications and attention in the visual modules, language-model
prefill, and 128 decoding steps, with a multiply-add counted as two FLOPs.

\subsection{Visual Budget and Cost}

\begin{table}[t]
\centering
\tablefont
\setlength{\tabcolsep}{5pt}
\renewcommand{\arraystretch}{1.05}
\caption{
Joint evidence and Bank budget.
Memory is peak GPU allocation. Higher Avg-9 and lower memory/TFLOPs are
better. \ranklegend
}
\label{tab:app_budget_cost}
\begin{tabular}{rccc}
\toprule
Tokens per stream & Avg-9 & Memory (GB) & TFLOPs \\
\midrule
32  & 0.742 & \best{15.72}  & \best{2.580} \\
64  & 0.760 & \second{15.83}  & \second{2.920} \\
\rowcolor{OursRowBlue}
\textbf{96}  & 0.769 & 15.95 & 3.304 \\
128 & \second{0.771} & 16.06  & 3.580 \\
192 & \best{0.772} & 16.27 & 4.150 \\
\bottomrule
\end{tabular}
\end{table}

From 96 to 192 tokens per stream, Avg-9 increases by only 0.003 while peak
memory rises by 2.0\% and TFLOPs by 25.6\%.
The 96-token setting therefore captures most of the observed gain with a
lower computational cost and nearly unchanged memory use. Other visual
interface studies combine token pruning and merging
\citep{shang2025llava,zhong2025aim}, or use context sparsification and
visual token pruning \citep{huang2025dynamic,ye2025fit}.
These approaches motivate distinguishing token count from the information
retained at a given budget.

\subsection{Model-Level Visual Interfaces}

\begin{table*}[t]
\centering
\tablefont
\setlength{\tabcolsep}{5pt}
\renewcommand{\arraystretch}{1.05}
\caption{
Implementation-level visual-token and computational profiles under the
profiling protocol above. ``All'' indicates that all retained foreground
patch features are processed before the compact language interface. ``N/A''
denotes a quantity that is unavailable or not comparable for the source
interface; it is not treated as zero. Lower cost is better. \ranklegend
}
\label{tab:app_model_efficiency}
\adjustbox{max width=\textwidth}{%
\begin{tabular}{lccccc}
\toprule
Method
& \shortstack{Input Patch\\Tokens}
& \shortstack{Context\\Latents}
& \shortstack{LLM Visual\\Tokens}
& \shortstack{Peak Memory\\(GB)}
& TFLOPs \\
\midrule
GPT-4o          & N/A    & N/A           & N/A    & N/A    & N/A \\
Quilt-LLaVA     & 576    & N/A           & 576    & 16.35 & 10.20 \\
SlideChat       & 20,480 & N/A           & 20,480 & 18.90 & 269.80 \\
WSI-LLaVA       & All    & Slide encoder & 576    & 17.35 & 10.49 \\
WSI-VQA         & All    & All           & N/A    & \best{6.80}  & \best{1.60} \\
MLLM-HWSI       & \shortstack{All}
                & 21
                & 1,941
                & 18.60
                & 28.40 \\
HistoSelect     & \shortstack{All}
                & N/A
                & 5,000
                & 19.10
                & 67.80 \\
\rowcolor{OursRowBlue}
\textbf{\method}& All    & 96            & \best{96}     & \second{15.95}  & \second{3.304} \\
\bottomrule
\end{tabular}%
}
\end{table*}

\method examines all retained foreground features in Preserve and Focus, but
Qwen receives only 96 fused evidence anchors. Under this
profile, it uses 15.95~GB of peak memory and 3.304~TFLOPs. Relative to
SlideChat, the implementation uses fewer language-side visual tokens and
substantially lower measured TFLOPs under the same profiling setup.

\subsection{Scaling with Foreground Patch Count}

\begin{figure*}[t]
\centering
\includegraphics[width=0.84\textwidth]{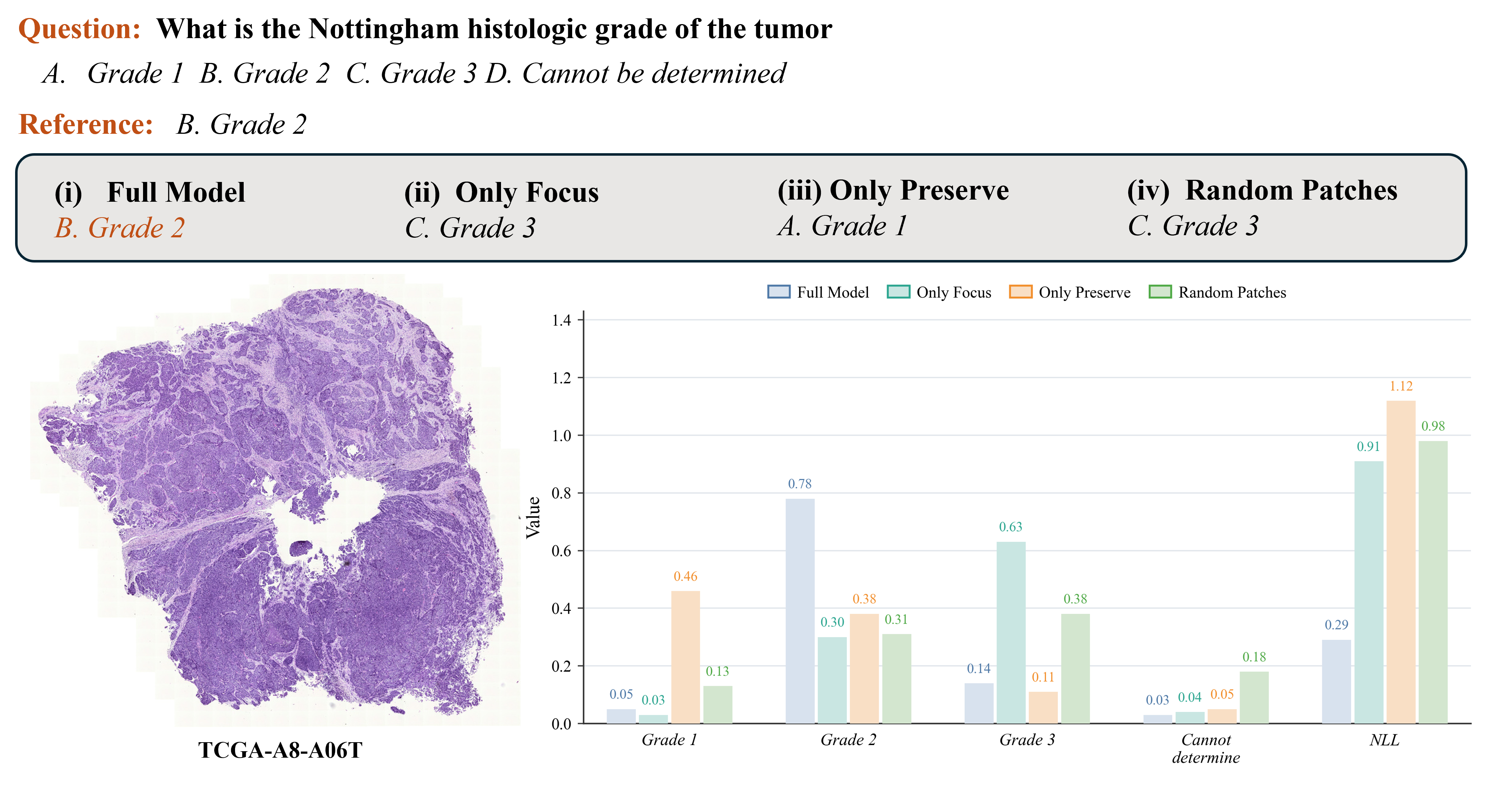}\\[0.2em]
\includegraphics[width=0.84\textwidth]{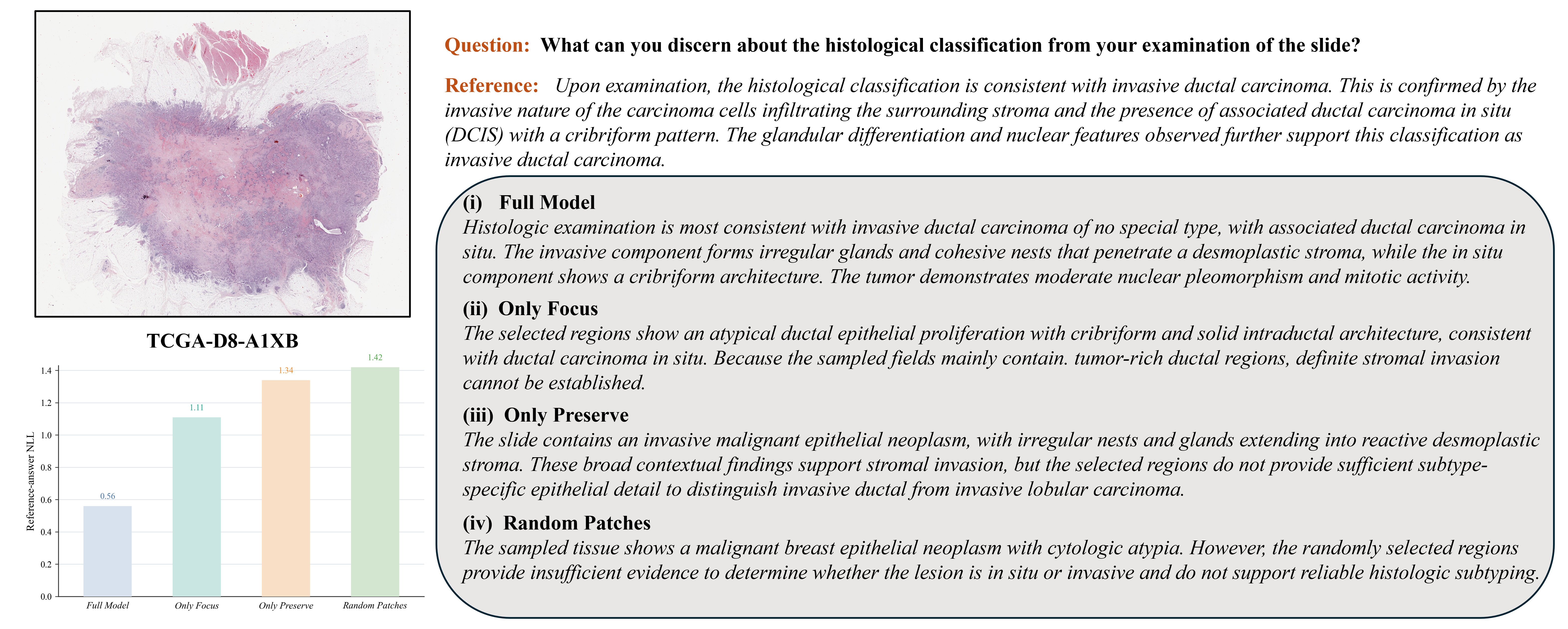}
\caption{
Qualitative evidence for complementary preservation and retrieval.
(a) Case 3 asks for the Nottingham histologic grade. The full model selects
Grade 2, whereas Focus-only and random-patch variants favor Grade 3 and
Preserve-only favors Grade 1; the full model also has the lowest displayed NLL.
(b) Case 4 asks for the histological classification. The full model gives
an invasive ductal carcinoma interpretation and cites both invasive glands
in desmoplastic stroma and an associated cribriform DCIS component.
}
\label{fig:app_case_studies}
\end{figure*}

\begin{table*}[t]
\centering
\tablefont
\setlength{\tabcolsep}{5pt}
\renewcommand{\arraystretch}{1.05}
\caption{
Single-question inference latency as the number of retained foreground
patches increases.
Times are in seconds with 95\% intervals and exclude CONCH feature
extraction. Lower latency is better. \ranklegend
}
\label{tab:app_latency_scaling}
\begin{tabular}{rcccc}
\toprule
& \multicolumn{2}{c}{\method}
& \multicolumn{2}{c}{SlideChat} \\
\cmidrule(lr){2-3}\cmidrule(lr){4-5}
$N_s$
& Time & 95\% interval
& Time & 95\% interval \\
\midrule
1,024  & \second{2.05} & [1.96, 2.17] & \best{1.68}  & [1.60, 1.80] \\
2,048  & \second{2.07} & [1.98, 2.19] & \best{1.94}  & [1.83, 2.08] \\
4,096  & \best{2.11} & [2.01, 2.24] & 2.47  & [2.31, 2.67] \\
8,192  & \best{2.20} & [2.08, 2.34] & 3.55  & [3.31, 3.84] \\
16,384 & \best{2.36} & [2.22, 2.52] & 5.92  & [5.50, 6.43] \\
32,768 & \best{2.65} & [2.48, 2.86] & 10.72 & [9.92, 11.70] \\
\bottomrule
\end{tabular}
\end{table*}

Long-context methods such as LongNet extend sequence processing through
sparse attention \citep{ding2023longnet}. \method keeps the downstream
visual sequence fixed, while slide-wide scoring remains linear in the
foreground patch count. Its measured latency therefore grows gradually
over the tested range.
At 32,768 patches, \method requires 2.65 seconds, compared with 10.72 for
SlideChat.

% Compact the closing analysis note so the limitations do not leave an orphan page.
\setlength{\parskip}{0.15\baselineskip}
\titlespacing*{\section}{0pc}{2ex plus 2pt minus 2pt}{5pt}
\section{Qualitative Analysis}
\label{app:qualitative}

Case 3 illustrates how broad review and focused examination complement one
another for a fine-grained grading decision. The reference answer is Grade 2, which is
also the full model prediction. Its displayed Grade 2 score is 0.78,
whereas the Focus-only, Preserve-only, and random-patch variants assign their
largest grade score to Grade 3, Grade 1, and Grade 3, respectively. The full
model also reports the smallest NLL value (0.29, compared with 0.91, 1.12,
and 0.98), consistent with assigning higher likelihood to the reference
answer under the reported scoring procedure. Combining the two views corrects
the errors made by either phase alone in this case.

Case 4 highlights the same Preserve--Focus complementarity for open-ended
classification.
The reference identifies invasive ductal carcinoma.  The full model
describes irregular glands and cohesive nests penetrating desmoplastic
stroma, together with a cribriform DCIS component and supporting nuclear
features. The Focus-only output recognizes ductal carcinoma in situ but
explicitly lacks evidence for stromal invasion; the Preserve-only output
recognizes invasion but cannot distinguish invasive ductal from invasive
lobular carcinoma; and random patches provide insufficient evidence to
distinguish in-situ from invasive disease or to identify the subtype. The lower
full-model NLL in the accompanying plot (0.56 versus 1.11, 1.34, and 1.42)
matches this qualitative difference in answer specificity: the fused answer
uses the Focus evidence together with the Morphology Bank.

\begin{minipage}{\linewidth}
\small
\setlength{\abovedisplayskip}{6pt}
\setlength{\belowdisplayskip}{6pt}
For a reference answer $Y=(y_1,\ldots,y_L)$, we define the displayed NLL
as the teacher-forced, length-normalized negative log-likelihood
\begin{equation}
\operatorname{NLL}(Y)=-\frac{1}{L}\sum_{t=1}^{L}
\log p_\theta(y_t\mid V,q,y_{<t}),
\label{eq:app_nll}
\end{equation}
using natural logarithms and excluding visual, question, boundary, and EOS
tokens from the average. For grading, the candidate strings ``Grade 1'',
``Grade 2'', and ``Grade 3'' are scored by
$u_j=\exp[-\operatorname{NLL}(Y_j)]/\sum_{r=1}^{3}
\exp[-\operatorname{NLL}(Y_r)]$.
These normalized candidate scores are not calibrated class probabilities.
\end{minipage}

\section{Limitations}
\label{app:limitations}
Our evaluation uses retrospective public cohorts, with external transfer
assessed only on SlideBench-BCNB.
The current implementation is limited to one CONCH--Qwen configuration
and a single $20\times$ scale.
All reported training and evaluation runs use random seed 42; we do not
report multi-seed performance variance or confidence intervals for predictive
metrics. The latency intervals in Appendix~\ref{app:efficiency} describe
repeated timing measurements.
Since both Preserve and Focus selection are learned from slide-level language
supervision, the selected patches are not pathologist-verified localizations.
Although the two displayed cases were reviewed by pathologists, selections
across the full test set have not been systematically validated against
expert-annotated regions. The resulting correspondence is functional, not a
claim of clinical equivalence to expert review.
Finally, Focus scans all retained foreground patches per question, so the
96-token fused prefix reduces decoder context rather than slide-wide retrieval
cost; Focus is compact at the language interface but not a
substitute for efficient slide indexing.
\end{document}